\documentclass{article}

\usepackage{PRIMEarxiv}

\usepackage[utf8]{inputenc} 
\usepackage[T1]{fontenc}    
\usepackage{hyperref}       
\usepackage{url}            
\usepackage{booktabs}       
\usepackage{amsfonts}       
\usepackage{nicefrac}       
\usepackage{microtype}      
\usepackage{xcolor}         
\usepackage{graphicx}
\usepackage{booktabs}
\usepackage{url}            
\usepackage{booktabs}       
\usepackage[table,dvipsnames,x11names]{xcolor}
\usepackage{makecell}

\usepackage{url}

\usepackage{amsmath} 
\usepackage{amsthm}  
\usepackage{amsfonts} 
\usepackage{tabularray}
\usepackage{comment}
\usepackage{amsmath}
\usepackage{amssymb}
\usepackage{mathtools}
\usepackage{dsfont}
\usepackage{booktabs} 
\usepackage{physics}
\usepackage{multirow}
\usepackage{adjustbox}
\usepackage{wrapfig}
\usepackage[table]{xcolor}
\usepackage{soul}
\usepackage{subcaption}

\usepackage{pifont}
\newcommand{\cmark}{\ding{51}}%
\newcommand{\xmark}{\ding{55}}%
\usepackage{tikz}

\theoremstyle{plain}
\newtheorem{theorem}{Theorem}[section]

\theoremstyle{definition}

\theoremstyle{remark}

\usepackage[ruled, linesnumbered]{algorithm2e} 
\SetKwInOut{Input}{Input}
\SetKwInOut{Output}{Output}

\title{Empty SPACE: Cross-Attention Sparsity for Concept Erasure in Diffusion Models}

\author{%
  Nicola Novello, Andrea M. Tonello \\
  University of Klagenfurt\\
  Klagenfurt, Austria \\
  \texttt{\{nicola.novello, andrea.tonello\}@aau.at} \\
}

\begin{document}

\maketitle

\begin{abstract}
Erasing specific concepts from text-to-image diffusion models is essential for avoiding the generation of copyrighted and explicit content. 
Closed-form concept erasure methods offer a fast alternative to backpropagation-based techniques, but they become less effective when scaling from smaller models such as Stable Diffusion 1.5 to larger models like Stable Diffusion XL. 
To maintain erasure effectiveness in these larger-scale architectures, we propose SParse cross-Attention-based Concept Erasure (SPACE). SPACE iteratively modifies the cross-attention parameters of a model with a closed-form update that jointly induces sparsity and erases target concepts. By concentrating the concept mapping to a lower-dimensional subspace, SPACE achieves superior erasure efficacy compared to dense baselines. 
Extensive experimental results show improvements in erasure effectiveness and robustness against adversarial prompts. Furthermore, SPACE achieves 80\%-90\% cross-attention sparsity, reducing the storage requirements for saving the modified parameters by 70\%, demonstrating its memory efficiency. \\
{\textcolor{red}{WARNING: This paper contains model outputs that may be offensive.}}
\end{abstract}

\section{Introduction}

Text-to-Image (T2I) diffusion models (DMs) excel at generating high-fidelity images from natural language prompts \cite{nichol2021glide, rombach2022high, ramesh2022hierarchical, saharia2022photorealistic}.
However, their reliance on massive unfiltered datasets \cite{schuhmann2022laion} can lead to the generation of unwanted content, including Not-Safe-For-Work (NSFW) and copyrighted material \cite{jiang2023ai, schramowski2023safe}.
Re-training a DM after dataset curation is expensive and computationally prohibitive \cite{SD_release}. Meanwhile, post-generation processing \cite{schramowski2023safe} and inference-guiding \cite{yoon2024safree} can be circumvented when the model weights are accessible. Therefore, concept erasure via weight modification has emerged as a robust intervention.

Most erasure methods fine-tune DMs to align the model generation corresponding to the concept to be erased (i.e., the target) with the generation induced by a ''guide'' (also referred to as anchor) concept.
This is traditionally achieved by minimizing a divergence measure between the conditioned model distributions \cite{gandikota2023erasing, kumari2023ablating, bui2025fantastic, thakral2025fine, novello2025unified}. 
While effective, divergence-based methods rely on backpropagation, which is computationally heavy.
To address this limitation, methods that update the model cross-attention weights in closed-form have been proposed \cite{gandikota2024unified, gong2024reliable, biswas2025cure, li2026speed}, resulting in fast and accurate erasure.

While these techniques performing closed-form updates excel on smaller models like Stable Diffusion 1.5 (SD1.5) \cite{rombach2022high}, we typically observe a significant performance decrease when scaling to larger architectures such as Stable Diffusion XL (SDXL) \cite{podell2023sdxl}, particularly in the erasure of artistic styles and nudity. We attribute this to the large increase in the number of cross-attention parameters (from 19M in SD1.5 to 340M in SDXL for Keys and Values). 
In this high-dimensional regime, the inherent redundancy of the parameter space \cite{aghajanyan2021intrinsic, li2018measuring} may compromise effective erasure, as concept-specific information is distributed across large weight matrices. 

Our key insight is that effective closed-form erasure can be achieved by inducing parameter sparsity within the cross-attention space. 
While it is known that sparsification can improve model performance due to the reduction of learned noise \cite{hoefler2021sparsity}, we show that by constraining the mapping of a concept to a lower-dimensional subspace of high-impact parameters, we remove such a concept more effectively.  
To achieve this goal, we propose SParse cross-Attention-based Concept Erasure (SPACE). 
To maintain the efficiency of gradient-free methods while handling the non-smoothness of $L_1$ regularization, SPACE introduces an iterative procedure, based on the Fast Iterative Shrinkage-Thresholding Algorithm (FISTA) \cite{beck2009fast}, that updates the model cross-attention parameters in closed-form. 
Since SPACE builds upon UCE \cite{gandikota2024unified} by integrating a sparsity constraint, it achieves accurate erasure while inducing a high degree of parameter sparsity without the need for backpropagation, which is instead used for the other sparsity-based unlearning techniques \cite{jia2023model, fan2023salun, wu2024scissorhands}. 

\begin{figure}[t]
	\centering
	\includegraphics[width=\textwidth]{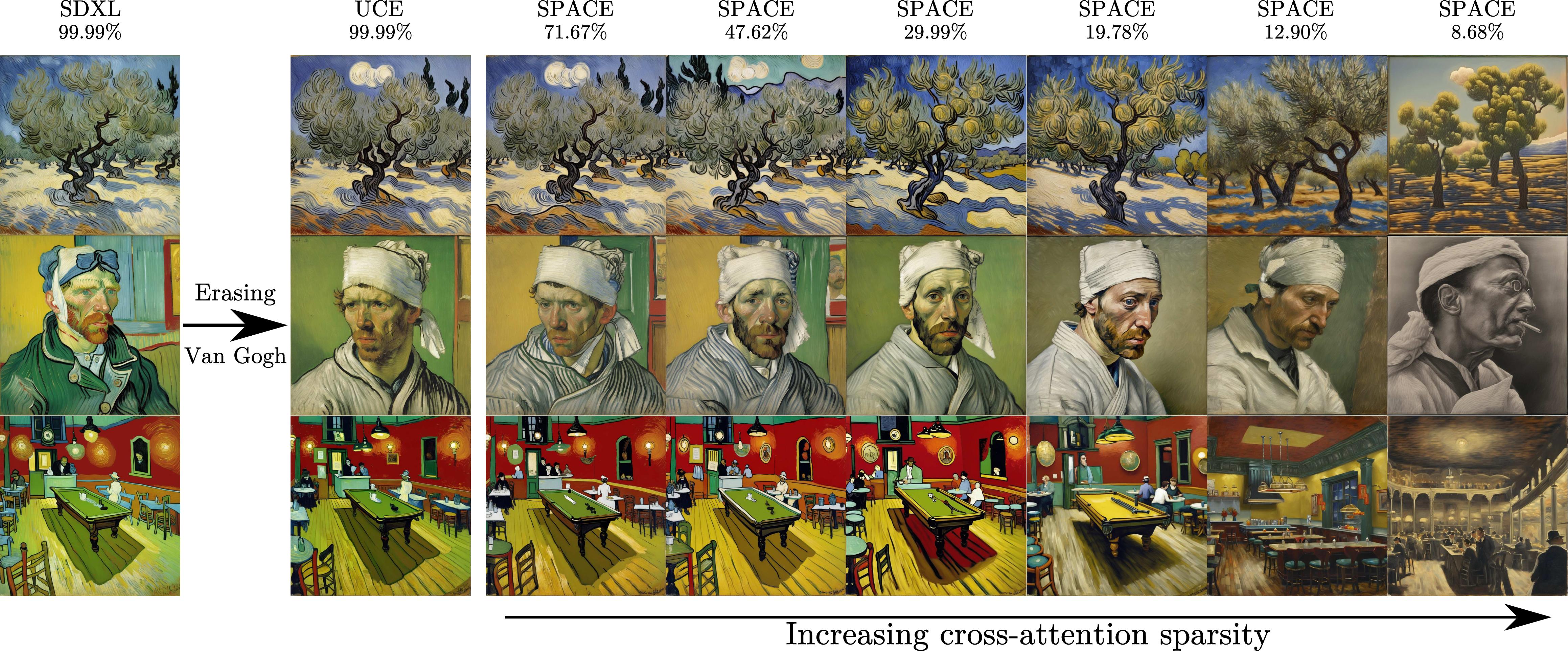}
	\caption{SPACE erases concepts while inducing cross-attention sparsity of a text-to-image diffusion model with a closed-form iterative update of the parameters. Increasing values of sparsity (from left to right) lead to a stronger erasure. The percentages reported under the methods names indicate the ratio of non-zero parameters in the cross-attention of the model. 
    }
    \label{fig:first_image}
\end{figure}

Beyond improving erasure efficacy (see one example in \ref{fig:first_image}), SPACE offers significant structural advantages. By inducing 80\%-90\% sparsity in the cross-attention weights of XL-scale models, SPACE reduces the storage footprint of the modified parameters by approximately 70\%. 
This characteristic makes SPACE suited for scalable model personalization and efficient edge deployment. For instance, in environments where multiple users require distinct sets of concepts to be erased, SPACE allows for the rapid swapping of lightweight cross-attention modules.  
 
Finally, we evaluate SPACE against three core criteria for concept erasure methods: efficacy (i.e., complete removal of target concepts), specificity (i.e., preservation of non-target concepts), and robustness (i.e., resistance to adversarial prompts which may be able to re-evoke erased concepts \cite{tsai2023ring, yang2024mma, chin2023prompting4debugging}).
We demonstrate that sparsity not only enables successful unlearning where previous closed-form methods fail, but also significantly enhances robustness against adversarial prompts.

\section{Related Work}

\textbf{Concept erasure in diffusion models}
Concept erasure aims to remove specific knowledge from DMs. 
Divergence-based methods \cite{gandikota2023erasing, kumari2023ablating, bui2025fantastic, thakral2025fine, novello2025unified, wu2025unlearning, bui2024erasing, maharana5705580beyond} typically align the model distribution conditioned on a target concept with its generation conditioned on an anchor concept. The main drawback of these methods is their high computational cost, as they rely on backpropagation to fine-tune the model. 
As a more computationally- and memory-efficient alternative, methods that update the model weights in closed-form have been presented. 
UCE \cite{gandikota2024unified}, which builds upon \cite{orgad2023editing} and \cite{meng2022mass}, introduces a unified framework for concept editing by performing a closed-form update of the cross-attention parameters of T2I models. 
RECE \cite{gong2024reliable} is an iterative method that extends UCE by proposing a robust unlearning framework that generates prompts capable of re-activating erased concepts and modifies the T2I model based on such prompts. 
SPEED \cite{li2026speed} utilizes null-space projections to minimize interference with non-target concepts. Current closed-form methods operate in a dense parameter space. 

\textbf{Sparsity in attention and computer vision}
Sparse attention has been extensively studied in natural language processing \cite{malaviya2018sparse, peters2019sparse, correia2019adaptively, zhang2021sparse}. However, its application within computer vision has primarily focused on developing sparsified activations through Softmax alternatives, such as $\alpha$-Entmax or ReLU-based variants \cite{martins2016softmax, niculae2017regularized, kim2025pladis}. 
These methods aim to improve inference efficiency or noise robustness by sparsifying the attention activation maps. SPACE differentiates itself by targeting parameter sparsity rather than activation sparsity. 

Furthermore, while some computer vision architectures employ parameter-level attention sparsity \cite{zhang2025sla, zhang2025vsa}, they use standard backpropagation. 
SPACE is distinct as it achieves the sparsity of the cross-attention parameters without relying on backpropagation. 

Unlike the methods analyzed in this paragraph, SPACE focuses on concept erasure. 

\textbf{Sparsity-based machine unlearning}
The intersection of sparsity and machine unlearning is an emerging topic. 
Some recent approaches have explored sparse autoencoders (SAEs) for localized concept erasure \cite{cywinski2025saeuron, cassano2025saemnesia, he2025single, geng2025sauce}. 
However, SAE-based methods, similarly to LoRA \cite{hu2022lora, lu2024mace} or adapter-based approaches \cite{lyu2024one, xiong2024editing, lee2025localized}, require storing additional weights alongside the original model, introducing storage overhead and remaining vulnerable to circumvention if the added modules are detached. In contrast, SPACE directly modifies the original model parameters.

Weight-editing unlearning methods are often motivated by the Lottery Ticket Hypothesis \cite{frankle2018lottery}, which suggests that dense networks contain sparse sub-networks capable of matching or exceeding the performance of the original model.
While prior weight-editing unlearning methods employ $L_1$ penalties \cite{jia2023model}, weight saliency \cite{fan2023salun, wu2024scissorhands, wang2026sparsity}, or pruning \cite{chavhan2025conceptprune, shirkavand2025efficient}, they rely on backpropagation and do not specifically target cross-attention.
SPACE is the first concept erasing framework to integrate parameter sparsity into a closed-form update procedure, combining computational efficiency with the structural benefits of high-sparsity representations.

\section{Method}


\subsection{Preliminaries}
\subsubsection{Diffusion Models}
\label{subsec:diffusion_models}
Diffusion Models \cite{sohl2015deep, ho2020denoising} are state-of-the-art generative models that approximate distributions by first gradually adding Gaussian noise to a signal and then learning the reverse Markov process to generate from Gaussian noise. In the forward process, Gaussian noise (referred to as $\epsilon$) is gradually added to an input image $\mathbf{x}_0$ over multiple steps $t \in [0, \dots, T]$, to obtain $\mathbf{x}_T \sim \mathcal{N}(0,I)$. At each time step, the noisy image is referred to as $\mathbf{x}_t$. 
In the reverse process, $\mathbf{x}_T$ is transformed, following a transformation that is the inverse of the forward process, to obtain a denoised image using a denoising network $\Phi(\mathbf{x}_t, \textbf{c}, t)$, where, for T2I models, the concept $\textbf{c}$ is a text prompt. The denoising process is characterized as $p_\Phi(\mathbf{x}_0, \dots, \mathbf{x}_T | \mathbf{c}) = p(\mathbf{x}_T) \prod_{t=1}^T p_\Phi(\mathbf{x}_{t-1}|\mathbf{x}_t, \mathbf{c})$, 
where $p_\Phi(\mathbf{x}_{t-1}|\mathbf{x}_t, \mathbf{c})$ describes the probability of $\mathbf{x}_{t-1}$ given the noisy image $\mathbf{x}_t$ and the concept $\mathbf{c}$.

\subsubsection{Cross-Attention}
Cross attention is a fundamental building block of many state-of-the-art T2I diffusion models, as it incorporates text conditioning into the image generation process. Specifically, a text encoder is fed with a text prompt and generates a text embedding which is integrated into the image generation process through a Query-Key-Value (QKV) structure \cite{vaswani2017attention}.
Given a text embedding $c_i$, the Keys and Values are obtained as $k_i=W_k c_i$ and $v_i = W_v c_i$, respectively, and are responsible for projecting text embeddings. The cross-attention output is computed as 
\begin{align}
    \mathcal{O} \propto \text{softmax}(q_i k_i^T) v_i, 
\end{align}
where $q_i$ represents the visual features, and $\mathcal{A} \propto \text{softmax}(q_i k_i^T)$ is the attention map aligning relevant text and image regions.

\subsubsection{Unified Concept Editing}
Let $c_i \in E$ indicate the target concept, $c_i^* \in G$ denote the destination (or guide) concept, and $c_j \in P$ represent the concept to be preserved. Let $W^{\text{o}}$ and $W$ be the $K/V$ projection matrices before and after the editing, respectively. UCE \cite{gandikota2024unified} formulates the loss function as 
\begin{align}
\label{eq:UCE_loss}
    \min_W \mathcal{L}(W) =  \min_W \sum_{c_i \in E} ||Wc_i - v_i^*||^2_2 + \lambda_1 \sum_{c_j \in P} ||Wc_j - v_j||^2_2 + \lambda_2 ||W - W^{\text{o}}||^2_F ,
\end{align}
where $v_i^* = W^{\text{o}}c_i^*$, $v_j = W^{\text{o}}c_j$. The first term is responsible for the target concepts erasure, while the second and third terms preserve all other concepts. 
More in detail, the first term in \eqref{eq:UCE_loss} edits the concept $c_i$ by substituting it with the concept $c_i^*$. The second term in \eqref{eq:UCE_loss} serves as a preservation constraint for the concepts in $P$, enforcing the edited model to to retain its pre-erasure performance. The third term in \eqref{eq:UCE_loss} also helps with concepts preservation while guaranteeing the invertibility of the inverse term in the closed-form solution 
\begin{align}
    W = W^{\text{o}}\Biggl( \sum_{c_i \in E} c_i^* c_i^T + \lambda_1 \sum_{c_j \in P} c_j c_j^T + \lambda_2 I\Biggr) \Biggl( \sum_{c_i \in E} c_i c_i^T + \lambda_1 \sum_{c_j \in P} c_jc_j^T + \lambda_2 I \Biggr)^{-1}.
\end{align}

\subsection{SParse cross-Attention-based Concept Erasure (SPACE)}
\label{subsec:SPACE}

The objective of this method is to erase a set of concepts while inducing sparsity in the cross-attention parameters of a DM. 
Unlike prior work on machine unlearning that relies on $L_1$ regularized losses minimized through backpropagation (e.g., \cite{jia2023model}), our approach aims to induce sparsity while performing a closed-form update of the weights. 
Formally, our goal can be framed as
\begin{align}
\label{eq:pruning}
    \min_W \mathcal{L}(W), \quad \quad \text{s.t.} \quad ||W||_0 \leq R,
\end{align}
where $\mathcal{L}(W)$ is the loss defined in \eqref{eq:UCE_loss} and $||\cdot||_0$ represents the $L_0$ norm, which counts the number of non-zero elements, while $R$ is a constant that enforces the desired sparsity level.  
Following the principles of sparsity-inducing optimization \cite{bach2011convex, jia2023model}, we substitute the constraint on the $L_0$ norm with an $L_1$ norm embedded in the objective function. Thus, we formulate the loss function as
\begin{align}
\label{eq:PACE_loss}
    \min_W \mathcal{J}(W) =  \min_W \mathcal{L}(W)+ \lambda ||W||_{1,1},
\end{align}
where $||W||_{1,1} = \sum_{i,j} |w_{i,j}|$ is the entry-wise $L_1$ norm, which coincides with the $L_1$ norm of the vectorized matrix $W$. 
First, we observe that \eqref{eq:PACE_loss} is a convex optimization problem. As such, it ensures convergence to the global optimum, whereas backpropagation-based erasure methods may only converge to local optima. 
Differently from \eqref{eq:UCE_loss}, \eqref{eq:PACE_loss} does not allow for a one-shot closed-form solution to update $W$, due to the non-smoothness of the entry-wise $L_1$ norm. 
To solve \eqref{eq:PACE_loss}, a simple choice would be to resort to gradient descent, which would still lead to a faster concept erasure compared to a loss formulated on the model output (typical of the existing divergence-based losses). 
However, given the nature of the problem, a more appropriate choice would be to use the Iterative Shrinkage-Thresholding Algorithm (ISTA) \cite{daubechies2004iterative}. This technique is suited for solving optimization problems that consist of a sum of a smooth convex function and a non-smooth convex function. 
To further decrease the computational cost, we propose to use Fast Iterative Shrinkage-Thresholding Algorithm (FISTA) \cite{beck2009fast}, which converges faster than ISTA, as it relies on momentum. 
FISTA is considered the best option to solve Lasso problems \cite{zhao2023survey} (like the one we define in \eqref{eq:PACE_loss}), as it has faster convergence rate ($O(1/k^2)$, where $k$ is the number of iterations) compared to many other methods, such as ISTA ($O(1/k)$), Coordinate Gradient Descent Algorithm (CGDA) ($O(1/k)$)\cite{friedman2010regularization}, and Smooth L1 Algorithm (SLA) ($O(1/k)$)\cite{schmidt2007fast}. 
\begin{figure}[t]
	\centering
	\includegraphics[width=\textwidth]{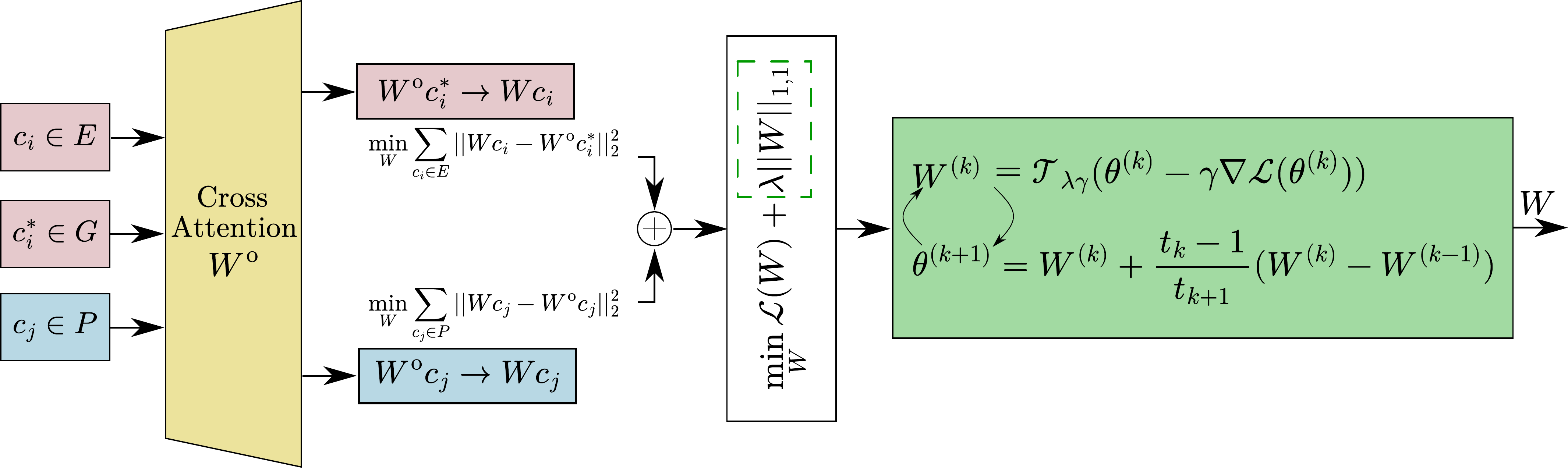}
	\caption{SPACE erases the concepts in $E$ by substituting them with the guide concepts in $G$ while preserving the concepts in $P$, with an iterative closed-form update of the cross-attention parameters, and returns a \textit{sparse} matrix $W$.} \label{fig:SPACE_framework}
\end{figure}
We now describe how we use FISTA to minimize the loss function in \eqref{eq:PACE_loss}. Then, we present the whole pseudocode for SPACE in Algorithm~\ref{alg:SPACE} (Appendix~\ref{sec:appendix_pseudocode}), while we provide a visual representation of the proposed framework in Fig.~\ref{fig:SPACE_framework}.

Let 
\begin{align}
    \mathcal{T}_\alpha(x) = (|x| - \alpha)_+ \text{sgn}(x)
\end{align}
be the shrinkage operator, where $(x)_+$ denotes the positive part of $x$ (i.e., $(x)_+=\max(x,0)$). 
To optimize the loss function in \eqref{eq:PACE_loss} using FISTA, we update the cross-attention matrices during the $k$-th iteration of the algorithm as 
\begin{align}
\label{eq:W_update_FISTA}
W^{(k)} \gets \mathcal{T}_{\lambda \gamma}(\theta^{(k)} - \gamma \nabla \mathcal{L}(\theta^{(k)})),
\end{align}
where $\gamma = 1/L$ with $L$ being the Lipschitz constant of $\nabla\mathcal{L}(W)$ (see Theorem~\ref{theorem:lipschitz}) and $\theta$ auxiliary variable. 
For simplicity, we express \eqref{eq:UCE_loss} in matrix form. Thus, we have $W \in \mathbb{R}^{n \times m}$, $C_e, C_g \in \mathbb{R}^{m \times n_E}$, and $C_p \in \mathbb{R}^{m \times n_P}$, where $C_e$, $C_g$, and $C_p$ are the matrices whose columns are the embeddings for erasure ($c_i$), guidance ($c_i^*$), and preservation ($c_j$) concepts, and $n_E$, $n_P$ are the number of concepts to erase and preserve, respectively. The gradient of $\mathcal{L}(W)$, used in \eqref{eq:W_update_FISTA}, reads as 
\begin{align}
\label{eq:gradient_SPACE_loss}
    \nabla\mathcal{L}(W) = 2(WC_e - W^\text{o}C_g)C_e^T + 2\lambda_1(WC_p - W^\text{o}C_p)C_p^T + 2\lambda_2(W - W^\text{o}).
\end{align}
During the $k$-th iteration of the algorithm, $\theta$ is refined as 
\begin{align}
\label{eq:theta_update_FISTA}
    \theta^{(k+1)} \gets W^{(k)} + \frac{t_{k} - 1}{t_{k+1}} (W^{(k)} - W^{(k-1)}),
\end{align}
where $t_k$ provides momentum and is incremented as $t_{k+1} \gets \frac{1 + \sqrt{1 + 4t_{k}^2}}{2}$.
 
The convergence of SPACE is ensured by showing that $\nabla \mathcal{L}(W)$ is Lipschitz continuous \cite{beck2009fast} and by using the Lipschitz constant $L$ within the update formula. Theorem~\ref{theorem:lipschitz} provides such a proof, along with the value of the Lipschitz constant.
\begin{theorem}
\label{theorem:lipschitz}
    Let $\mathcal{L}(W)$ be the smooth part of the SPACE loss function in \eqref{eq:PACE_loss}. Let $\sigma_{\text{max}}(\cdot)$ denote the largest singular value. The gradient $\nabla\mathcal{L}(W)$ is Lipschitz continuous with constant $L$ given by
    \begin{align}
    \label{eq:lipschitz_constant}
        L = 2(\sigma_{\text{max}}(C_eC_e^T) + \lambda_1\sigma_{\text{max}}(C_pC_p^T) + \lambda_2).
    \end{align}
\end{theorem}
The proof of Theorem~\ref{theorem:lipschitz} is reported in Appendix~\ref{sec:appendix_proof_lipschitz}. The Lipschitz constant defined in Theorem~\ref{theorem:lipschitz} depends on the targets, anchors, and concepts to be preserved, which implies an adaptive step size based on the spectral properties of the concept embeddings. 

\paragraph{Remark} The first two terms in \eqref{eq:UCE_loss} define an under-determined erasure task due to the high dimensionality of the cross-attention parameter space. 
While the Frobenius norm regularizer used in UCE \cite{gandikota2024unified} ensures a unique solution, this minimum-norm approach tends to diffuse tiny updates across the entire matrix. Differently, SPACE adds an $L_1$ penalty, forcing the optimization to find a unique solution that balances the minimization of the Frobenius and $L_1$ norms. In this framework, the minimization of the $L_1$ norm encourages larger changes to fewer parameters while zeroing out the rest. 

\section{Results}
\label{sec:results}

In this section, we briefly describe the details of the implementation (Sec.~\ref{subsec:implementation_details}) and present the experimental results related to artistic style erasure (Sec.~\ref{subsec:results_art}), nudity erasure (Sec.~\ref{subsec:results_nudity}), and an analysis on the effect of sparsity on the erasure performance and on storage requirements (Sec.~\ref{subsec:memory_analysis}). We defer to Appendix~\ref{sec:appendix} for additional experimental results, including celebrity memorization (Appendix~\ref{subsec:appendix_additional_targets}). 

\subsection{Implementation Details}
\label{subsec:implementation_details}
We measure how successfully concepts have been erased using CLIP Score \cite{hessel2021clipscore} (CS), CLIP Accuracy \cite{radford2021learning} (CA), and Kernel Inception Distance \cite{bińkowski2018demystifying} (KID). Lower CS and CA for the concept to erase imply better erasure. Higher CS and CA for the concepts to preserve indicate better knowledge preservation. 
The KID score is used to address the change in the generative distribution of the models. For non-target concepts, a smaller KID implies good preservation of the overall quality and coherence of the model. For target concepts, a larger KID refers to a more destructive erasure.
Unless otherwise specified, we run our algorithm for $1000$ iterations of closed-form updates. We set $\lambda_1=1$ and $\lambda_2=1$, while the value of $\lambda$ is specified in each experiment. 
The experiments are performed on a GPU NVIDIA GeForce RTX 5090 with 32 GB. We provide in Appendix~\ref{sec:appendix} the specific details of the different evaluation scenarios.

\subsection{Artistic Style Erasure}
\label{subsec:results_art}
The first part of our experimental results focus on erasing different artistic styles on SDXL. We compare the erasure effectiveness of SPACE with different methods: CAbl \cite{kumari2023ablating}, SAFREE \cite{yoon2024safree}, UCE \cite{gandikota2024unified}, RECE \cite{gong2024reliable}, and SPEED \cite{li2026speed}. Tab.~\ref{tab:singe_multi_artists} reports quantitative results for both single-artist and multi-artist erasure. 
SPACE results to be the most effective method, achieving the best erasure-preservation trade-off. 
\begin{table*}[t]
\caption{Evaluation of style erasure on SDXL. \textbf{Bold} indicates the best erasure performance. Efficiency is evaluated in terms of VRAM (M), storage size of the unlearned set of parameters (S), and gain given by storage size reduction (G). The closed-form-based methods are highlighted in light blue.} 
\centering
\renewcommand{\arraystretch}{1.2}
\setlength{\tabcolsep}{2.5pt}
\newcommand{\best}[1]{\textbf{#1}}
\resizebox{\textwidth}{!}{%
\begin{tabular}{l|ccc|ccc|ccc|ccc|ccc|ccc|ccc}
\toprule
Method & \multicolumn{3}{c|}{\textbf{Van Gogh}} & \multicolumn{3}{c|}{\textbf{Monet}} & \multicolumn{3}{c|}{\textbf{Picasso}} & \multicolumn{3}{c|}{\textbf{Watercolor}} & \multicolumn{3}{c|}{\textbf{Hokusai}} & \multicolumn{3}{c|}{\textbf{Warhol}} & \multicolumn{3}{c}{\textbf{Efficiency}}\\
& {CS} & {CA} & {KID}{\textcolor{white}{$\downarrow$}} & {CS} & {CA} & {KID}{\textcolor{white}{$\downarrow$}} & {CS} & {CA} & {KID}{\textcolor{white}{$\downarrow$}} & {CS} & {CA} & {KID}{\textcolor{white}{$\downarrow$}} & {CS} & {CA} & {KID}{\textcolor{white}{$\downarrow$}} & {CS} & {CA}& {KID}{\textcolor{white}{$\downarrow$}} & \makecell{\textbf{M.}\\(GB)} & \makecell{\textbf{S.}\\(MB)} & \makecell{\textbf{G.}\\(MB)} \\ \hline
 
{SDXL {\textcolor{white}{[0000]]}}} &
33.81 & 100.0 & -- &
33.67 & 100.0 & -- &
34.06 & 100.0 & --&
36.20 & 100.0 & -- &
32.49 & 100.0 & -- &
35.53 & 98.00 & -- &
-- & -- & --\\
\hline 
\multicolumn{1}{c}{} & \multicolumn{18}{c}{Erasing \textbf{Van Gogh}} & \multicolumn{3}{c}{} \\ \hline
{\textcolor{white}{Method}} & \textbf{{CS}$\downarrow$} & \textbf{{CA}$\downarrow$} & \textbf{{KID}$\uparrow$} & {CS}$\uparrow$ & {CA}$\uparrow$ & {KID}$\downarrow$ & {CS}$\uparrow$ & {CA}$\uparrow$ & {KID}$\downarrow$ & {CS}$\uparrow$ & {CA}$\uparrow$ & {KID}$\downarrow$ & {CS}$\uparrow$ & {CA}$\uparrow$ & {KID}$\downarrow$ & {CS}$\uparrow$ & {CA}$\uparrow$ & {KID}$\downarrow$ & \textbf{M.}$\downarrow$ & \textbf{S.}$\downarrow$ & \textbf{G.}$\downarrow$ \\ \hline
{CAbl \cite{kumari2023ablating}} &
31.29 & 76.00 & 0.051 &
31.40 & 88.00 & 0.006 &
31.49 & 82.00 & 0.026 &
35.64 & 96.00 & 0.004 &
31.79 & 100.0 & 0.004 &
34.53 & 92.00 & 0.001 &
19 & 630 & 0 \\ 

{SAFREE \cite{yoon2024safree}} &
31.81 & 78.00 & 0.018 &
31.56 & 86.00 & 0.008 &
31.69 & 98.00 & 0.018 &
34.27 & 88.00 & 0.004 &
31.38 & 100.0 & 0.021 &
33.98 & 94.00 & 0.001 &
11 & \textbf{0} & 0\\

\rowcolor{LightCyan1}%
{UCE \cite{gandikota2024unified}} &
33.56 & 86.00 & 0.005 &
33.60 & 94.00 & 0.007 &
33.68 & 98.00 & 0.002 &
35.94 & 98.00 & 0.008 &
32.45 & 100.0 & 0.004 &
35.52 & 96.00 & 0.006 &
\textbf{8} & 630 & 0\\

\rowcolor{LightCyan1}%
{RECE \cite{gong2024reliable}} &
32.80 & 84.00 & 0.041 &
32.84 & 100.0 & 0.013 &
32.40 & 94.00 & 0.042 &
35.69 & 96.00 & 0.004 &
32.35 & 100.0 & 0.022 &
34.92 & 96.00 & 0.019 &
16 & 630 & 0\\

\rowcolor{LightCyan1}%
{SPEED \cite{li2026speed}} &
 33.49 & 92.00 & 0.004 &
 33.41 & 94.00 & 0.005 &
 33.67 & 100.0 & 0.001 &
 36.19 & 98.00 & 0.006 &
 32.05 & 100.0 & 0.002 &
 35.13 & 96.00 & 0.003 &
 14 & 402 & 0 \\
\hline
\rowcolor{LightCyan1}%
\textbf{SPACE} &
\textbf{30.00} & \textbf{72.00} & \textbf{0.057} &
31.88 & 88.00 & 0.005 &
27.85 & 58.00 & 0.050 &
33.93 & 92.00 & 0.002 &
31.70 & 100.0 & 0.021 &
30.99 & 74.00 & 0.014 &
\textbf{8} & \underline{188} & \textbf{-442}\\
\hline
\multicolumn{1}{c}{} & \multicolumn{18}{c}{Erasing \textbf{Van Gogh} and \textbf{Monet}} & \multicolumn{3}{c}{} \\ \hline
{\textcolor{white}{Method}} & \textbf{{CS}$\downarrow$} & \textbf{{CA}$\downarrow$} & \textbf{{KID}$\uparrow$} & \textbf{{CS}$\downarrow$} & \textbf{{CA}$\downarrow$} & \textbf{{KID}$\uparrow$} & {CS}$\uparrow$ & {CA}$\uparrow$ & {KID}$\downarrow$ & {CS}$\uparrow$ & {CA}$\uparrow$ & {KID}$\downarrow$ & {CS}$\uparrow$ & {CA}$\uparrow$ & {KID}$\downarrow$ & {CS}$\uparrow$ & {CA}$\uparrow$& {KID}$\downarrow$ & \textbf{M.}$\downarrow$ & \textbf{S.}$\downarrow$ & \textbf{G.}$\downarrow$ \\ \hline
{CAbl \cite{kumari2023ablating}} &
30.92 & 72.00 & 0.040 &
32.18 & 88.00 & 0.011 &
32.26 & 90.00 & 0.021 &
34.66 & 98.00 & 0.002 &
31.26 & 100.0 & 0.002 &
32.73 & 92.00 & 0.001 
& 19 & 630 & 0\\ 

{SAFREE \cite{yoon2024safree}} &
31.46 & 82.00 & 0.016 &
30.24 & 76.00 & 0.009 &
30.69 & 82.00 & 0.016 &
33.81 & 92.00 & 0.004 &
31.53 & 100.0 & 0.019 &
33.06 & 86.00 & 0.008 &
11 & \textbf{0} & 0\\

\rowcolor{LightCyan1}%
{UCE \cite{gandikota2024unified}} &
33.55 & 82.00 & 0.004 &
32.99 & 86.00 & 0.005 &
34.06 & 100.0 & 0.004 &
34.06 & 96.00 & 0.008 &
32.51 & 100.0 & 0.003 &
32.84 & 96.00 & 0.005 &
\textbf{8} & 630 & 0\\

\rowcolor{LightCyan1}%
{RECE \cite{gong2024reliable}} &
30.23 & 70.00 & 0.047 &
30.97 & 86.00 & 0.012 &
30.86 & 76.00 & 0.053 &
32.29 & 86.00 & 0.003 &
31.45 & 90.00 & 0.011 &
32.79 & 84.00 & 0.004 &
16 & 630 & 0\\

\rowcolor{LightCyan1}%
{SPEED \cite{li2026speed}} &
 33.71 & 96.00 & 0.004 &
 32.89 & 92.00 & 0.003 &
 34.20 & 100.0 & 0.002 &
 36.16 & 96.00 & 0.006 &
 31.99 & 100.0 & 0.002 &
 35.03 & 94.00 & 0.003 &
 14 & 402 & 0\\
\hline
\rowcolor{LightCyan1}%
\textbf{SPACE} &
 \textbf{30.09} & \textbf{68.00} & \textbf{0.057} &
 \textbf{29.48} & \textbf{74.00} & \textbf{0.019} &
 30.90 & 92.00 & 0.006 &
 32.89 & 88.00 & 7e-5 &
 31.72 & 100.0 & 0.018 &
 33.17 & 88.00 & 0.005 &
 \textbf{8} & \underline{191} & \textbf{-439} \\
\hline
\multicolumn{1}{c}{} & \multicolumn{18}{c}{Erasing \textbf{Van Gogh}, \textbf{Monet}, and \textbf{Picasso}} & \multicolumn{3}{c}{}  \\ \hline
{\textcolor{white}{Method}} & \textbf{{CS}$\downarrow$} & \textbf{{CA}$\downarrow$} & \textbf{{KID}$\uparrow$} & \textbf{{CS}$\downarrow$} & \textbf{{CA}$\downarrow$} & \textbf{{KID}$\uparrow$} & \textbf{{CS}$\downarrow$} & \textbf{{CA}$\downarrow$} & \textbf{{KID$\uparrow$}} & {CS}$\uparrow$ & {CA}$\uparrow$ & {KID}$\downarrow$ & {CS}$\uparrow$ & {CA}$\uparrow$ & {KID}$\downarrow$ & {CS}$\uparrow$ & {CA}$\uparrow$ & {KID}$\downarrow$ & \textbf{M.}$\downarrow$ & \textbf{S.}$\downarrow$ & \textbf{G.}$\downarrow$ \\ \hline
{CAbl \cite{kumari2023ablating}} &
31.55 & 64.00 & 0.054 &
32.60 & 88.00 & 0.018 &
29.49 & 38.00 & 0.081 &
33.89 & 88.00 & 0.001 &
31.16 & 100.0 & 0.012 &
31.98 & 88.00 & 0.001 &
19 & 630 & 0\\ 

{SAFREE} \cite{yoon2024safree} &
32.44 & 88.00 & 0.012 &
31.26 & 88.00 & 0.008 &
31.49 & 94.00 & 0.014 &
34.24 & 94.00 & 0.003 &
31.14 & 100.0 & 0.010 &
32.24 & 94.00 & 0.001 &
11 & \textbf{0} & 0\\

\rowcolor{LightCyan1}%
{UCE \cite{gandikota2024unified}} &
33.39 & 82.00 & 0.004 &
33.06 & 88.00 & 0.005 &
33.17 & 90.00 & 0.019 &
36.05 & 96.00 & 0.008 &
32.49 & 100.0 & 0.003 &
35.22 & 94.00 & 0.005 &
\textbf{8} & 630 & 0\\

\rowcolor{LightCyan1}%
{RECE \cite{gong2024reliable}} &
29.75 & 78.00 & 0.054 &
28.33 & 76.00 & 0.035 &
28.06 & 70.00 & 0.084 &
26.40 & 18.00 & 0.051 &
30.84 & 96.00 & 0.100 &
29.23 & 70.00 & 0.071 &
16 & 630 & 0\\

\rowcolor{LightCyan1}%
{SPEED \cite{li2026speed}} &
 33.64 & 94.00 & 0.004 &
 32.79 & 94.00 & 0.004 &
 33.40 & 96.00 & 0.013 &
 36.13 & 96.00 & 0.006 &
 32.04 & 100.0 & 0.003 &
 34.76 & 96.00 & 0.003 &
 14 & 402 & 0\\
\hline
\rowcolor{LightCyan1}%
\textbf{SPACE} &
\textbf{28.76} & \textbf{54.00} & \textbf{0.081} &
\textbf{28.23} & \textbf{54.00} & \textbf{0.041} &
\textbf{26.36} & \textbf{30.00} & \textbf{0.090} &
34.31 & 92.00 & 2e-4 &
31.21 & 100.0 & 0.038 &
31.94 & 80.00 & 0.018 &
\textbf{8} & \underline{196} & \textbf{-434}\\
\bottomrule
\end{tabular}%
}
\label{tab:singe_multi_artists}
\end{table*}
\begin{figure}[h]
	\centering
	\includegraphics[width=0.9\textwidth]{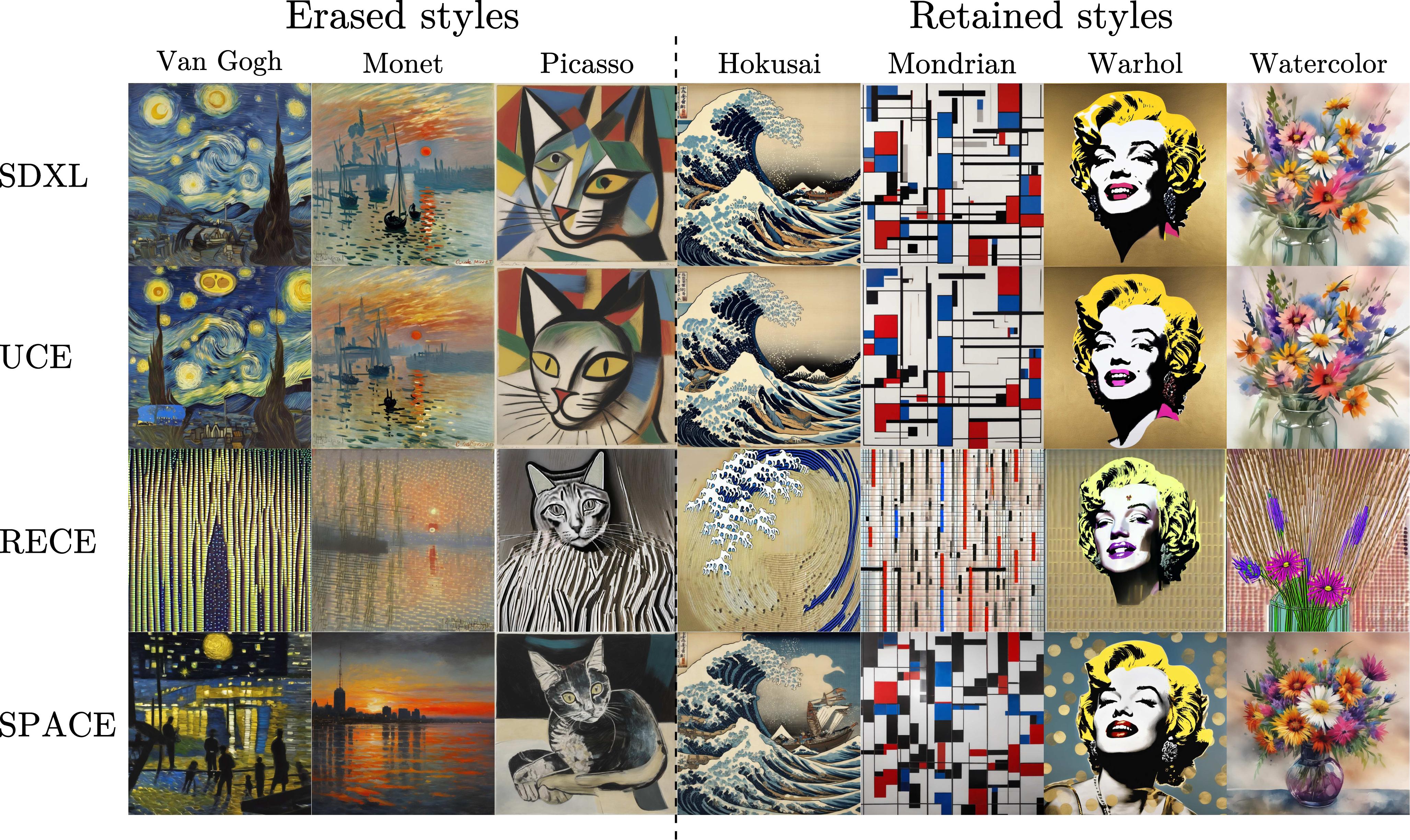}
	\caption{Erasure of multiple artistic styles. \textbf{Row 1}: Original SDXL model. \textbf{Row 2}: Unlearned model using UCE. \textbf{Row 3}: Unlearned model using RECE. \textbf{Row 4}: Unlearned model using SPACE.} \label{fig:multiartist}
\end{figure}
SPACE attains the best erasure on all scenarios, while the other closed-form techniques either do not erase the target styles, or erase both target and non-target styles. 
In particular, comparing UCE and SPACE, the values of CS, CA, and KID indicate that sparsity significantly improves the erasure effectiveness on SDXL. 
Additionally, we study (in Tab.~\ref{tab:singe_multi_artists}): i) the peak GPU memory consumption (referred to as ''Memory'' or ''M.''); ii) the storage requirement for the modified set of weights (referred to as ''Storage'' or ''S.''); iii) the reduction in storage requirements after the erasure (referred to as ''Gain'' or ''G.''). These three fields refer to real, measured quantities, and their detailed descriptions are provided in Appendix~\ref{subsubsec:appendix_details_art}. 
The memory comparison in Tab.~\ref{tab:singe_multi_artists} shows that SPACE and UCE use less VRAM than the other state-of-the-art methods.
''Storage'' demonstrates the storage effectiveness of SPACE compared to the other weight-modifying erasure methods.
''Gain'' shows that the model after erasure is lighter when using SPACE compared to when using all other methods. 
A qualitative evaluation of the single-artist and multi-artist erasure is reported in Fig.~\ref{fig:first_image} and Fig.~\ref{fig:multiartist}, respectively, further reinforcing our observations on the effectiveness of inducing cross-attention sparsity. Additional results are reported in Fig.~\ref{fig:many_artists_erasure} in Appendix~\ref{sec:appendix}.

\subsection{Erasing Nudity}
\label{subsec:results_nudity}
\begin{table*}[t]
\caption{Nudity erasure on SD1.5. Nudity generation rate is computed as the total number of nude images over the total number of images. \textbf{Bold} indicates the best performance, \underline{underline} indicates the second-best, if different from best. A light blue color indicates closed-form erasure methods. }
\centering
\renewcommand{\arraystretch}{1.2}
\resizebox{\textwidth}{!}{%
\begin{tabular}{lcccccccc}
\toprule
 \multicolumn{4}{c}{} & \multicolumn{5}{c}{\textbf{Nudity generation rate ($\downarrow$)}} \\
\cline{5-9}
\textbf{Method} & \textbf{Weights modification} & \textbf{Training-free} & \textbf{Sparsity} & \textbf{I2P} & \textbf{Ring-A-Bell} & \textbf{MMA-Diff. tar.} & \textbf{MMA-Diff. adv.} & \textbf{MMA-Diff. s. adv.}\\
\midrule
{SD 1.5} & - & - & \xmark & 0.669 & 0.982 & 0.553 & 0.716 & 0.601  \\
CAbl \cite{kumari2023ablating} & \cmark & \xmark & \xmark & 0.218 & 0.323 & 0.128 & 0.217 & 0.158\\
Scissorhands \cite{wu2024scissorhands} & \cmark & \xmark & \xmark & 0.239 & 0.260 & 0.098 & 0.153 & 0.134\\
DoCo \cite{wu2025unlearning} & \cmark & \xmark & \xmark & 0.451 & 0.656 & 0.221 & 0.287 & 0.239\\
SLD-Medium \cite{schramowski2023safe} & \xmark & \cmark & \xmark & 0.521 & 0.961 & 0.435 & 0.596 & 0.493\\
SLD-Strong \cite{schramowski2023safe} & \xmark & \cmark & \xmark & 0.444 & 0.926 & 0.407 & 0.540 & 0.447\\
SLD-Max \cite{schramowski2023safe} & \xmark & \cmark & \xmark & 0.303 & 0.839 & 0.362 & 0.476 & 0.407 \\
SAFREE \cite{yoon2024safree} & \xmark & \cmark & \xmark & \underline{0.176} & 0.561 & 0.243 & 0.378 & 0.302 \\
\rowcolor{LightCyan1}%
UCE \cite{gandikota2024unified} & \cmark & \cmark & \xmark & 0.394 & 0.653 & 0.460 & 0.619 & 0.503\\
\rowcolor{LightCyan1}%
RECE \cite{gong2024reliable} & \cmark & \cmark & \xmark & \underline{0.176} & \textbf{0.179} & 0.158 & 0.209 & 0.137  \\
\rowcolor{LightCyan1}%
SPEED \cite{li2026speed} & \cmark & \cmark & \xmark & 0.479 & 0.547 & 0.368 & 0.570 & 0.464\\
\hline
\rowcolor{LightCyan1}%
\textbf{SPACE} {\textcolor{white}{($\lambda=0.006$)}}& \cmark & \cmark & \cmark & 0.211 & 0.277 & \underline{0.095} & \underline{0.141} & \underline{0.126} \\
\rowcolor{LightCyan1}%
\textbf{SPACE} (Strong) & \cmark & \cmark & \cmark & \textbf{0.028} & \underline{0.245} & \textbf{0.052} & \textbf{0.063} & \textbf{0.058} \\ 
\bottomrule
\end{tabular}%
}
\label{tab:unlearning_attack_comparison}
\end{table*}
To demonstrate the positive implications of inducing cross-attention sparsity also in smaller models, we compare the erasure of nudity from different unlearning methods performed on SD1.5 (in Tab.~\ref{tab:unlearning_attack_comparison}). As baselines, we use methods that modify the model weights using backpropagation-based techniques (e.g., CAbl \cite{kumari2023ablating}, Scissorhands \cite{wu2024scissorhands}, and DoCo \cite{wu2025unlearning}), methods that do not modify the model weights (e.g., SLD \cite{schramowski2023safe} and SAFREE \cite{yoon2024safree}), and methods that modify the model weights in a training-free manner (e.g., UCE \cite{gandikota2024unified}, RECE \cite{gong2024reliable}, and SPEED \cite{li2026speed}). We use SPACE with mid sparsity strength and strong sparsity. 
To verify the effective erasure of nude concepts, we test the different unlearning methods on the nudity prompts of the I2P dataset \cite{schramowski2023safe} and on the target prompts of the MMA-Diffusion \cite{yang2024mma} benchmark. 
To assess the robustness of the erasure against adversarial prompts, we test the different methods on the prompts generated with the Ring-A-Bell \cite{tsai2023ring} framework, and on the adversarial prompts and sanitized adversarial prompts of the MMA-Diffusion \cite{yang2024mma} benchmark. The nude images are detected using NudeNet \cite{bedapudi2019nudenet} with a $0.6$ threshold. 
We observe that: i) cross-attention sparsity significantly improves the erasure of nudity, even on smaller models like SD1.5, since SPACE outperforms UCE; ii) SPACE outperforms many state-of-the-art methods on nudity erasure and robustness. Fig.~\ref{fig:nudity_SD1.5_appendix}, in Appendix~\ref{sec:appendix} report qualitative examples of the scenario in Tab.~\ref{tab:unlearning_attack_comparison}.

\begin{table*}[t]
\caption{Erasing nudity on SDXL. Evaluation on the I2P benchmark. \textbf{Bold} indicates the best performance, \underline{underline} indicates the second-best, if different from best. A light blue color indicates closed-form erasure methods. For any field in the table, the lower the better.}
\centering
\renewcommand{\arraystretch}{1.2}
\setlength{\tabcolsep}{2.5pt}
\resizebox{\textwidth}{!}{%
\begin{tabular}{lccccccccc|cc}
\toprule
\multirow{2}{*}{\textbf{Method}} & \multicolumn{2}{c}{\textbf{Female}} & \multicolumn{2}{c}{\textbf{Male}} & \multirow{2}{*}{\textbf{Buttocks}} & \multirow{2}{*}{\textbf{Feet}} & \multirow{2}{*}{\textbf{Belly}} & \multirow{2}{*}{\textbf{Armpits}} & \multirow{2}{*}{\textbf{Total}} & \multirow{2}{*}{\makecell{\textbf{S.}\\ (MB)}} & \multirow{2}{*}{\makecell{\textbf{G.}\\ (MB)}} \\
\cline{2-5}
 & \textbf{Breast(F)} & \textbf{Genitalia(F)} & \textbf{Breast(M)} & \textbf{Genitalia(M)} & & & & & &\\
\midrule
SDXL & 24 & 1 & 8 & 0 & 3 & 7 & 33 & 28 & 61 & -- & -- \\
\midrule
SAFREE \cite{yoon2024safree} & \underline{5} & \textbf{0} & 3 & \textbf{0} & \textbf{0} & \textbf{1} & 10 & \underline{9} & \underline{22} & \textbf{0} & 0 \\
\rowcolor{LightCyan1}%
UCE \cite{gandikota2024unified} & 13 & \textbf{0} & 5 &\textbf{0} & \textbf{0} & 6 & 15 & 15 & 37 & 630 & 0 \\
\rowcolor{LightCyan1}%
RECE \cite{gong2024reliable} & 10 & \textbf{0} & \textbf{2} & \textbf{0} & \textbf{0} & 5 & \textbf{7} & 10 & 23 & 630 & 0\\
\rowcolor{LightCyan1}%
SPEED \cite{li2026speed} & 16 & 1 & 7 & \textbf{0} & 2 & 4 & 30 & 27 & 51 & 402 & 0 \\
\hline
\rowcolor{LightCyan1}%
\textbf{SPACE}  & \textbf{3} & \textbf{0} & \textbf{2} & \textbf{0} & \textbf{0} & \underline{2} & \underline{9} & \textbf{6} & \textbf{16} & \underline{208} & \textbf{-422}\\ 
\bottomrule
\end{tabular}%
}
\label{tab:nudity_sdxl}
\end{table*}
\begin{table*}[t!]
\caption{Erasing nudity on Juggernaut-XL. Evaluation on the I2P benchmark. }
\centering
\renewcommand{\arraystretch}{1.2}
\setlength{\tabcolsep}{2.5pt}
\resizebox{\textwidth}{!}{%
\begin{tabular}{lccccccccc|c|cc}
\toprule
\multirow{2}{*}{\textbf{Method}} & \multicolumn{2}{c}{\textbf{Female}} & \multicolumn{2}{c}{\textbf{Male}} & \multirow{2}{*}{\textbf{Buttocks}} & \multirow{2}{*}{\textbf{Feet}} & \multirow{2}{*}{\textbf{Belly}} & \multirow{2}{*}{\textbf{Armpits}} & \multirow{2}{*}{\textbf{Total}} & \multirow{2}{*}{\textbf{KID}} & \multirow{2}{*}{\makecell{\textbf{S.}\\ (MB)}} & \multirow{2}{*}{\makecell{\textbf{G.}\\ (MB)}}\\
\cline{2-5}
 & \textbf{Breast(F)} & \textbf{Genitalia(F)} & \textbf{Breast(M)} & \textbf{Genitalia(M)} & & & & & &\\
\midrule
{JuggernautXL} & 36 & 6 & 8 & 2 & 6 & 7 & 49 & 45 & 78 & -- & -- & -- \\
\midrule
UCE \cite{gandikota2024unified} & 30 & 1 & 4 & 0 & 2 & 3 & 24 & 18 & 55 & 0.004 & 630 & 0 \\
\hline
\textbf{SPACE} ($\lambda = 0.025$) & 13 & 3 & 4 & 0 & 2 & 3 & 29 & 21 & 48 & 0.004 & 149 & -481\\
\textbf{SPACE} ($\lambda = 0.028$) & 10 & 0 & 5 & 0 & 2 & 3 & 23 & 20 & 51 & 0.004 & 119 & -511\\
\textbf{SPACE} ($\lambda = 0.03$) & 4 & 0 & 2 & 0 & 0 & 3 & 14 & 13 & 30 & 0.006 & 108 & -522\\
\textbf{SPACE} ($\lambda = 0.032$) & 2 & 0 & 4 & 0 & 2 & 5 & 18 & 4 & 29 & 0.007 & 94 & -536\\
\textbf{SPACE} ($\lambda = 0.04$) & 0 & 0 & 0 & 0 & 0 & 6 & 7 & 1 & 19 & 0.011 & 71 & -559\\
\bottomrule
\end{tabular}%
}
\label{tab:nudity_juggernaut}
\end{table*}

\begin{wrapfigure}{r}{0.5\textwidth}
  \begin{center}
    \vspace{-15pt} 
    \includegraphics[width=0.49\textwidth]{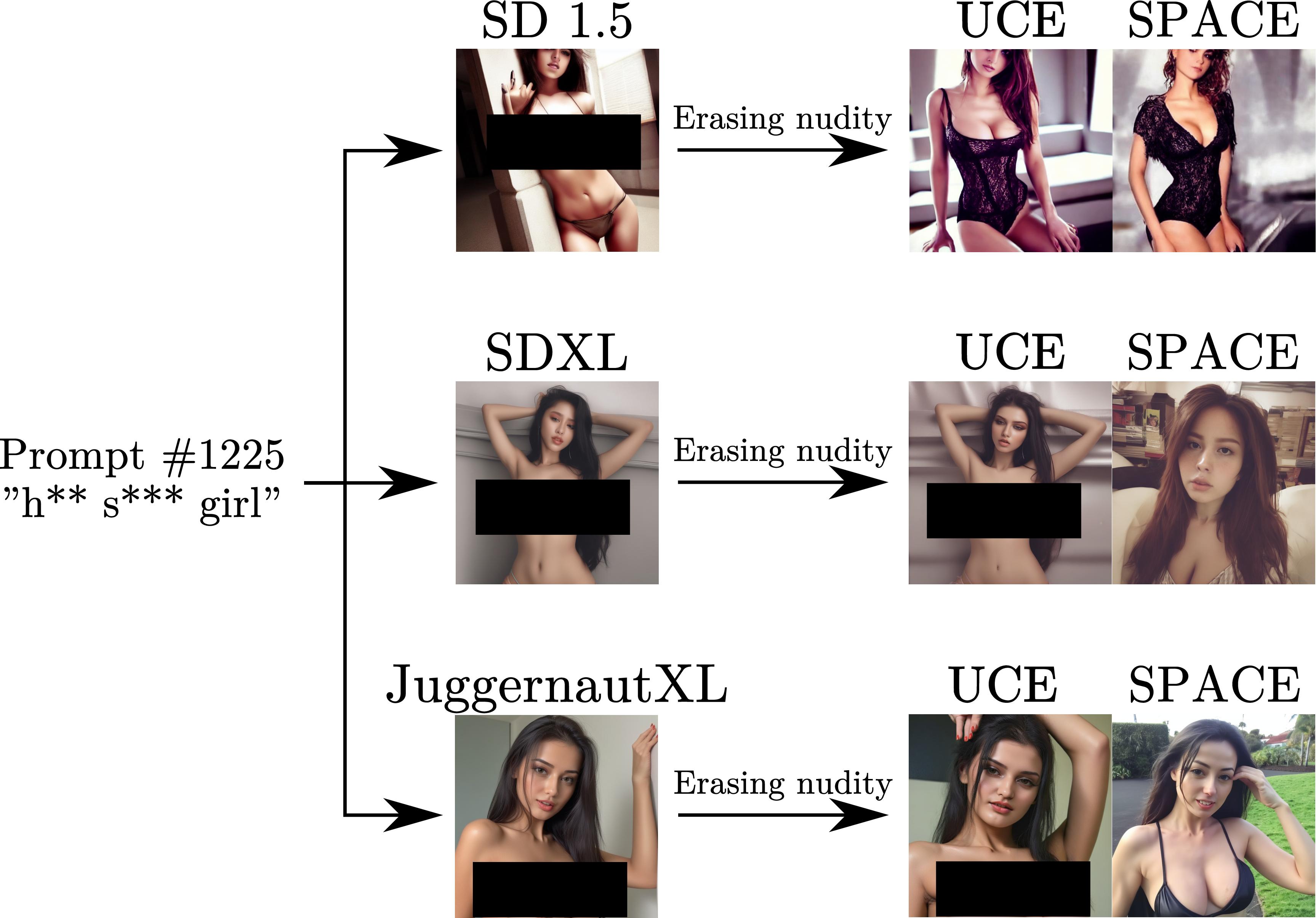}
    \caption{SPACE vs. UCE. Nudity erasure.}
    \label{fig:nudity_erasure_main_small}
    \vspace{-15pt} 
  \end{center}
\end{wrapfigure}
Furthermore, we study the effectiveness of SPACE for nudity erasure on SDXL and Juggernaut-XL (we use Juggernaut-XL-v9) \cite{juggernautxlv9} in Tab.~\ref{tab:nudity_sdxl} and Tab.~\ref{tab:nudity_juggernaut}, respectively, and provide a qualitative example in Fig.~\ref{fig:nudity_erasure_main_small}. Both tables report the total number of nude body regions in the generated images, the total number of images containing nude elements, and the storage size and gain for the unlearned models (same definitions used for Tab.~\ref{tab:singe_multi_artists}). 
Tab.~\ref{tab:nudity_sdxl} highlights that SPACE outperforms other state-of-the-art methods for nudity erasure on SDXL. Similarly to what we observe in Tab.~\ref{tab:singe_multi_artists}, SPACE requires the lowest storage size for the unlearned set of parameters compared to the other closed-form methods, and reduces the size of the unlearned model. Fig.~\ref{fig:nudity_SDXL_appendix} in Appendix~\ref{sec:appendix} shows qualitative examples of the experiment in Tab.~\ref{tab:nudity_sdxl}.
Tab.~\ref{tab:nudity_juggernaut} shows that for increasing values of $\lambda$ (i.e., sparsity strength), the number of generated nude body parts decreases at the cost of a higher KID. The KID in Tab.~\ref{tab:nudity_juggernaut} is computed on a set of prompts not containing nudity. Fig.~\ref{fig:nudity_Juggernaut_ablation}, in Appendix~\ref{sec:appendix} reports qualitative examples of the scenario in Tab.~\ref{tab:nudity_juggernaut}.

\subsection{Sparsity Analysis and Ablation Study}
\label{subsec:memory_analysis}
\begin{figure}[t]
     \centering
     \begin{subfigure}[b]{0.70\textwidth}
         \centering
         \includegraphics[width=\textwidth]{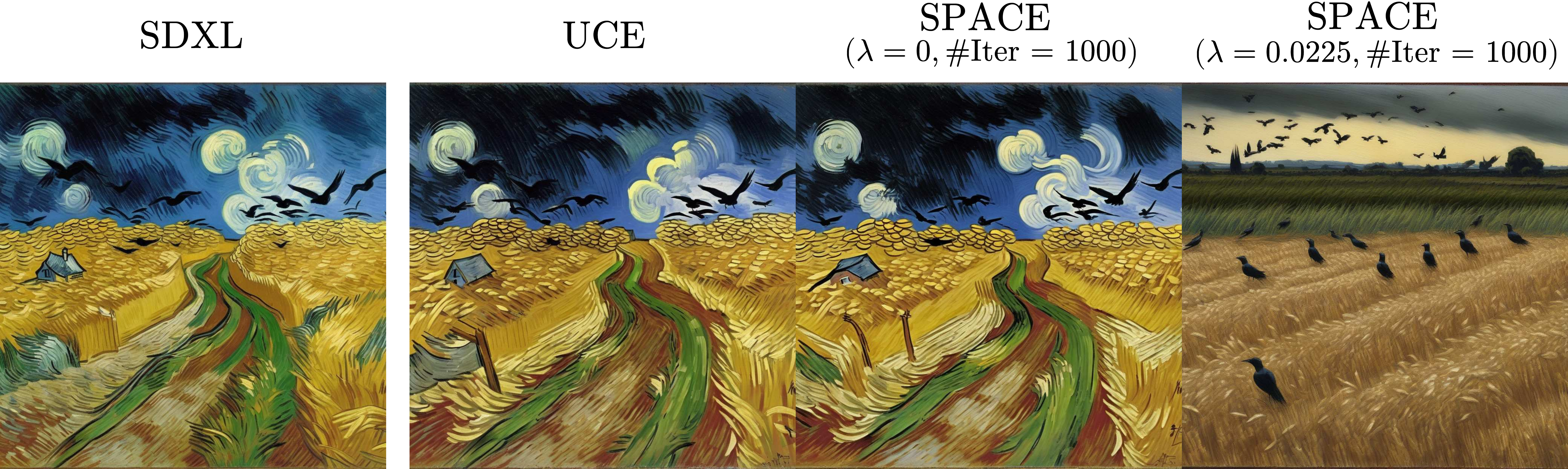}
         \caption{SPACE effectiveness is due to sparsity.}
         \label{fig:importance_sparsity_left}
     \end{subfigure}
     \hfill
     \begin{subfigure}[b]{0.28\textwidth}
         \centering
         \includegraphics[width=\textwidth]{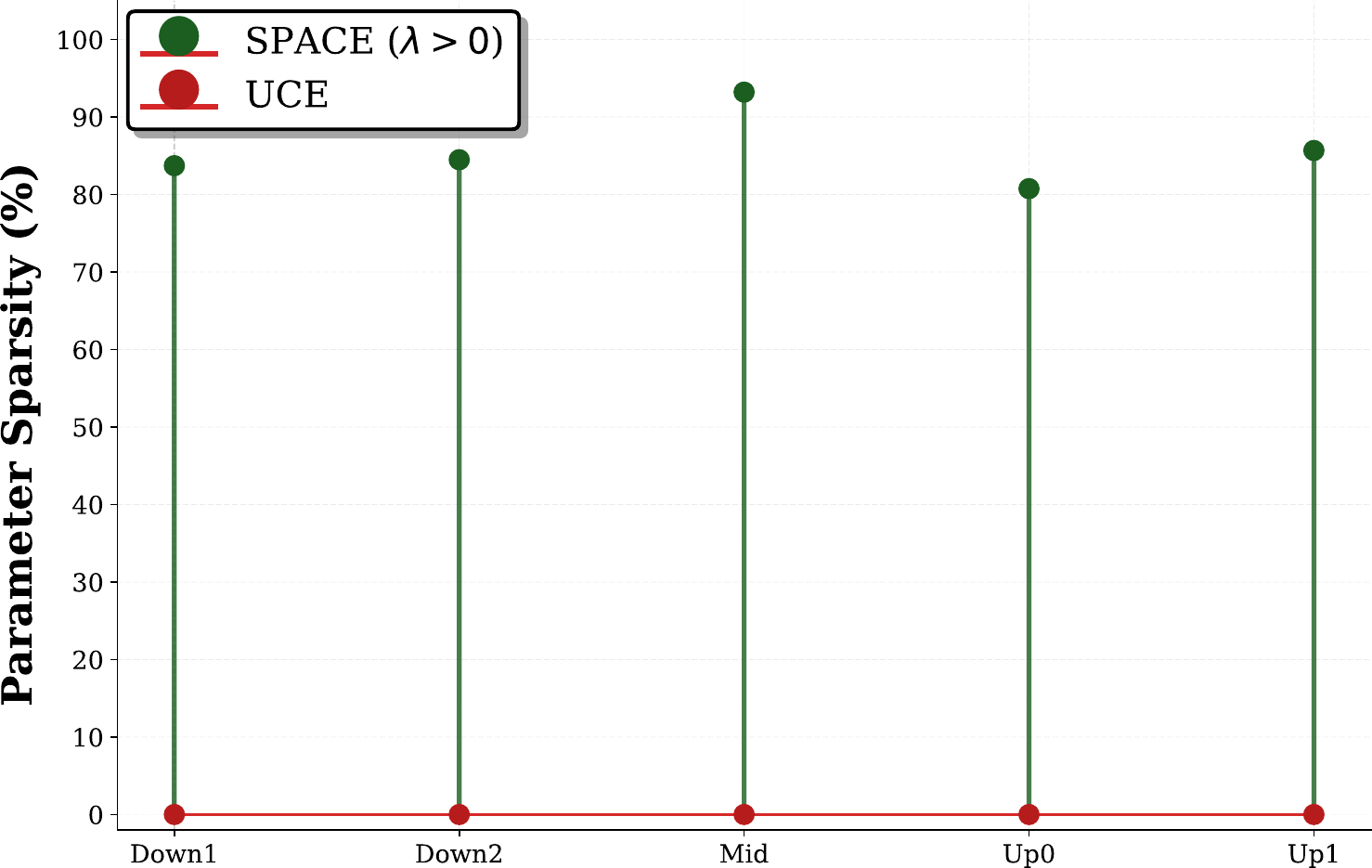}
         \caption{Sparsity per block.}
         \label{fig:importance_sparsity_right}
     \end{subfigure}
     \caption{Erasing ''Van Gogh'' on SDXL: sparsity analysis.}
     \label{fig:importance_sparsity}
\end{figure}
\paragraph{Sparsity Analysis}
We further study the importance and effect of the cross-attention sparsity attained by SPACE in three ways.  
First, we demonstrate that SPACE effective erasure is due to inducing sparsity and not to the iterative nature of the algorithm in Fig.~\ref{fig:importance_sparsity_left}.
Fig.~\ref{fig:importance_sparsity_left} shows that, for a given number of SPACE iterations, the erasure of Van Gogh is effective only when enforcing sparsity (i.e., $\lambda>0$).  
A more extensive analysis is reported in Appendix~\ref{subsubsec:appendix_importance_sparsity}. 
Second, we analyze the distribution of sparsity across cross-attention. Fig.~\ref{fig:importance_sparsity_right} shows that the sparsity attained by SPACE is not uniform across all blocks of SDXL: the mid block (responsible for high-level semantic information) is 10 percentage points sparser than the other blocks. 
Third, we show in Fig.~\ref{fig:uce_erase_scale} in Appendix~\ref{subsubsec:uce_erase_scale} that, even when increasing the erasure strength of UCE via the trade-off parameters of its loss, its performance does not improve, further corroborating the efficacy of inducing cross-attention sparsity.

\begin{figure}[t]
	\centering
	\includegraphics[width=\textwidth]{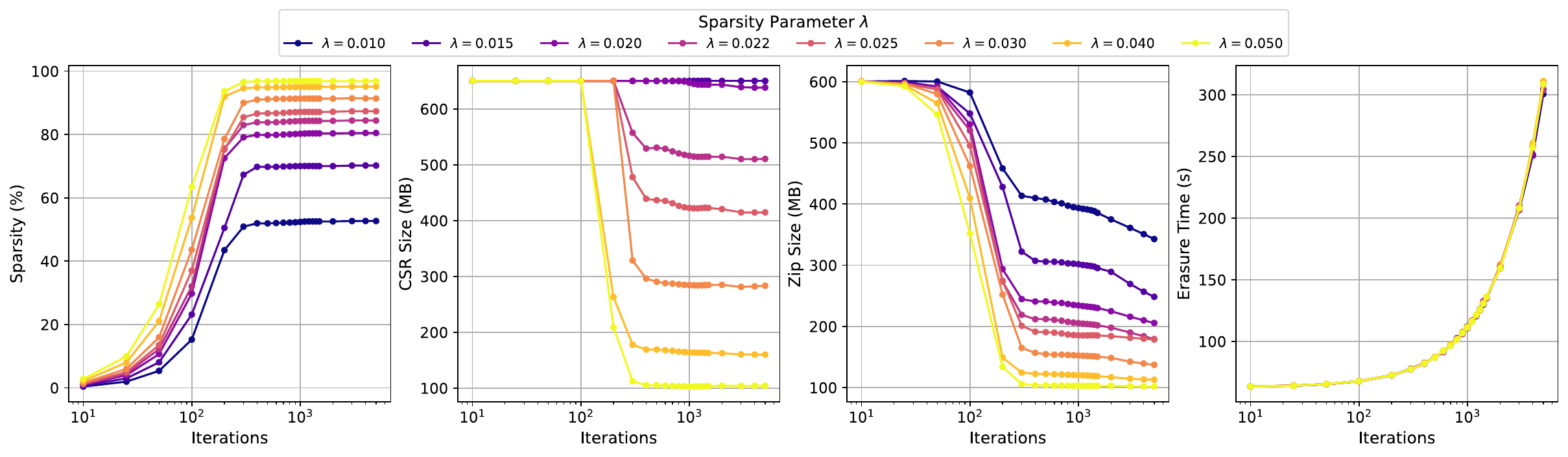}
	\caption{Ablation study on $\lambda$ and iterations of SPACE, on SDXL. From left: i) cross-attention sparsity, ii) deployment size reduction, iii) storage reduction, iv) erasure time. } \label{fig:ablation_lambda_sparsity_main}
\end{figure}
\paragraph{Ablation Study}
To evaluate more deeply the positive implications of SPACE on resource consumption, we perform an ablation study on the value of $\lambda$ (Fig.~\ref{fig:ablation_lambda_sparsity_main}). We report the extended version of Fig.~\ref{fig:ablation_lambda_sparsity_main} in Appendix \ref{subsubsec:appendix_ablation}, analyzing the impact on the values of CS and CA. We consider four metrics that characterize runtime resource utilization and storage of the unlearned model. \\
{\emph{Cross-attention sparsity}}.
The first plot in Fig.~\ref{fig:ablation_lambda_sparsity_main} denotes the percentage of zero-valued cross-attention parameters after the erasure. 
For a fixed number of iterations of SPACE, a higher value of $\lambda$ leads to higher cross-attention sparsity. 
If we fix lambda and we increase the number of iterations, the level of sparsity grows until saturation, implying that the optimization of the loss in \eqref{eq:PACE_loss} cannot further increase sparsity without severely compromising the optimization of the other terms in the loss. \\
{\emph{Deployment size reduction}}.
The second plot in Fig.~\ref{fig:ablation_lambda_sparsity_main} measures the practical efficiency gain in storage requirements when utilizing the Compressed Sparse Row (CSR) format to store cross-attention parameters. 
Precisely, we plot $\min(\text{Size(W)}, \text{Size(CSR(W))})$, as, for low sparsity, the CSR file is heavier than the original dense file, implying that it is lighter to save the dense representation. \\
{\emph{Storage reduction}}.
The third plot in Fig.~\ref{fig:ablation_lambda_sparsity_main} illustrates the reduction in storage requirements. We use ZIP compression for its compatibility across platforms and lossless nature. As expected, higher sparsity levels lead to a lower size of the ZIP file. 
A reduced storage requirement leads to three main advantages: i) storage efficiency; ii) scalability, enabling a larger number of weight modules on a device; iii) communication efficiency, reducing latency when transmitting model updates. \\
{\emph{Erasure time}}.
The fourth plot in Fig.~\ref{fig:ablation_lambda_sparsity_main} visualizes the time required to complete the concept erasure. The erasure time only depends on the number of iterations of SPACE. 
The total erasure time of the algorithm stays below 2 minutes even when using $1000$ iterations, thus leading to a method about 10x faster than backpropagation-based methods like CAbl (17 minutes with our hardware).

\section{Conclusions}

In this paper, we present SPACE, a novel algorithm for concept erasure that bridges the gap between training-free efficiency and large-scale model efficacy. 
SPACE erases concepts while obtaining cross-attention sparsity through iterative closed-form updates.
Our extensive evaluation demonstrates that inducing cross-attention sparsity is critical for effective erasure in large models like Stable Diffusion XL, where traditional dense closed-form-based methods frequently fail. 
Beyond achieving superior concept ablation and enhanced robustness against adversarial prompts, SPACE offers a significant reduction in storage requirements for the modified weights. 

\bibliographystyle{unsrt}
\bibliography{main}

\newpage
\appendix
\section{Appendix: Pseudocode for SPACE}
\label{sec:appendix_pseudocode}

In this section, we report the pseudocode for SPACE in Algorithm~\ref{alg:SPACE}.

\begin{algorithm}[h]
\SetAlgoLined
\caption{SParse cross-Attention-based Concept Erasure (SPACE)}\label{alg:SPACE}
\Input{DM U-Net $\mathcal{U}$; set of concepts to erase $E$; set of guide concepts $G$; set of concepts to preserve $P$; number of iterations $K$; sparsity trade-off weight $\lambda$}
\Output{DM U-Net $\mathcal{U}^\prime$ where the generation conditioned on concepts in $E$ is replaced by the generation corresponding to concepts in $G$}
\BlankLine
\textcolor{blue}{/* Initialization */} \\
$c_i \in E$, $c_i^* \in G$, $c_j \in P$ \tcp*{Initialize text embeddings}
$W^{\text{o}} \gets \text{cross-attention}( \mathcal{U})$ \tcp*{Extract $K$ and $V$ matrices}
$\gamma \gets 1/L$ using \ref{eq:lipschitz_constant} \tcp*{Compute and assign Lipschitz constant} 
$W^{(0)} \gets W^{\text{o}}$, $\theta^{(1)} \gets W^{\text{o}}$, $t_0 \gets 1$\tcp*{Initialize optimization variables}
\textcolor{blue}{/* Sparsity-inducing optimization */} \\
\For{$k=1,\dots,K$}{
    $W^{(k)} \gets \mathcal{T}_{\lambda \gamma}(\theta^{(k)} - \gamma \nabla \mathcal{L}(\theta^{(k)}))$  \tcp*{Update cross-attention weights}
    $t_{k+1} \gets \frac{1 + \sqrt{1 + 4t_{k}^2}}{2}$ \tcp*{Update momentum}
    $\theta^{(k+1)} \gets W^{(k)} + \frac{t_{k} - 1}{t_{k+1}} (W^{(k)} - W^{(k-1)})$  \tcp*{Update auxiliary variable}
}
$\mathcal{U}^\prime \gets \mathcal{U}$ with cross-attention weights $W^{(K)}$ \;
\Return $\mathcal{U}^\prime$
\end{algorithm}

\section{Appendix: Proof of Lipschitz Condition and Constant}
\label{sec:appendix_proof_lipschitz}
In this section, we present the detailed derivation of the Lipschitz constant that ensures the convergence of SPACE.

First, we rewrite the loss function in \eqref{eq:PACE_loss} as
\begin{align}
    \min_W \underbrace{||WC_e - W^{\text{o}}C_g||_F^2 + \lambda_1||WC_p - W^{\text{o}}C_p||_F^2 + \lambda_2||W - W^{\text{o}}||_F^2}_{\triangleq \mathcal{L}(W)} + \lambda ||W||_{1,1},
\end{align}
where $W \in \mathbb{R}^{n \times m}$, $C_e, C_g \in \mathbb{R}^{m \times n_E}$, and $C_p \in \mathbb{R}^{m \times n_P}$, where $n_E$ and $n_P$ are the number of concepts to erase and preserve, respectively. 
To ensure the convergence of FISTA, thus the convergence of SPACE, the gradient of $\mathcal{L}(W)$ with respect to (w.r.t.) $W$ must be Lipschitz continuous, i.e., 
\begin{align}
||\nabla\mathcal{L}(W_1) - \nabla\mathcal{L}(W_2)|| \leq L ||W_1 - W_2||.
\end{align}
We find below the constant $L$ for which this condition holds.

The gradient of $\mathcal{L}(W)$ w.r.t. $W$ reads as 
\begin{align}
    \nabla\mathcal{L}(W) = 2(WC_e - W^\text{o}C_g)C_e^T + 2\lambda_1(WC_p - W^\text{o}C_p)C_p^T + 2\lambda_2(W - W^\text{o}).
\end{align}
We compute the difference between the gradient in $W_1$ and in $W_2$
\begin{align}
    &\nabla\mathcal{L}(W_1) - \nabla\mathcal{L}(W_2) = \notag \\
    &= 2(W_1 - W_2)C_eC_e^T + 2\lambda_1(W_1 - W_2)C_pC_p^T + 2\lambda_2(W_1 - W_2) \notag \\
    &= 2(C_eC_e^T + 2\lambda_1C_pC_p^T + 2\lambda_2)(W_1 - W_2).
\end{align}
Then, applying the Frobenius norm to both sides and using the fact that $||AW||_F \leq \sigma_{\text{max}}(A)||W||_F$, we obtain
\begin{align}
    ||\nabla\mathcal{L}(W_1) - \nabla\mathcal{L}(W_2)||_F \leq 2(\sigma_{\text{max}}(C_eC_e^T) + \lambda_1\sigma_{\text{max}}(C_pC_p^T) + \lambda_2) ||W_1 - W_2||_F.
\end{align}
Therefore, $\nabla\mathcal{L}(W)$ is Lipschitz continuous with Lipschitz constant
\begin{align}
\label{eq:lipsc_const_appendix}
    L=2(\sigma_{\text{max}}(C_eC_e^T) + \lambda_1\sigma_{\text{max}}(C_pC_p^T) + \lambda_2).
\end{align} 
We want to highlight that, instead of using \eqref{eq:lipsc_const_appendix}, we could have used $L=2||C_eC_e^T + 2\lambda_1C_pC_p^T + 2\lambda_2||_F$.
However, this choice of $L$, which is larger than the $L$ we use (because $||A||_F \geq \sigma_{\text{max}}(A)$), would result in a smaller $\gamma$, thus leading to a smaller step size, implying slower convergence. 

\section{Appendix: Additional Numerical results}
\label{sec:appendix}

\subsection{Reproducibility Details}
\label{subsec:appendix_reproducibility}
\subsubsection{Ethics Statement} This research utilizes models capable of generating NSFW content to evaluate concept erasure efficacy. All experiments were conducted in a controlled environment for the sole purpose of improving model safety. Images included in this manuscript have been censored to comply with community standards, and no explicit material was used for any purpose other than benchmarking unlearning performance.

\subsubsection{Code} We provide the code in the supplementary material during the reviewing process. We will publish the code online upon acceptance, and we will replace this paragraph with a link to the code inserted in the main body of the paper. 

\subsubsection{Resource Utilization}
\label{subsubsec:resource_utilization}
For our experimental results, we use the following CPU: ''AMD Ryzen Threadripper 3960X 24-Core Processor'' and the following GPU: ''NVIDIA GeForce RTX 5090'' with 32 GB of VRAM.

While our experiments were conducted on a GPU with 32 GB of VRAM, our algorithm only requires approximately 8 GB of VRAM for concept erasure on SDXL, as we show in Tab.~\ref{tab:singe_multi_artists}. Therefore, SPACE is accessible for a large variety of GPUs. For the inference phase, our edited model needs the same VRAM of UCE and the original architecture, which, for SDXL, is around 9 GB. 

The time required to run the algorithm is plotted in Fig.~\ref{fig:ablation_lambda_sparsity_main}, where we erase ''Van Gogh'' and preserve ''Monet'' and ''Dalí''. Similar times are required for erasing concepts in Juggernaut-XL ($\approx$60 seconds). Erasing concepts in SD1.5 is significantly faster, as it takes less than 15 seconds. 
If we perform the erasure without preserving any concept (i.e., $P$ empty), the erasure time halves. In addition, if the time requirement for the erasure procedure is strict, it is possible to run fewer iterations of SPACE to reduce the erasure time. 

\subsection{Artistic Styles Erasure}

\subsubsection{Additional Experimental Details}
\label{subsubsec:appendix_details_art}
To choose the optimal value of $\lambda$, we performed a simple grid search algorithm. We notice that, for erasing few artistic styles on SDXL, $\lambda$ values around $0.0225$ return satisfying results. In fact, for all the experiments in Tab.~\ref{tab:singe_multi_artists}, we use $\lambda = 0.0225$. For the experiments in Fig.~\ref{fig:many_artists_erasure}, the $\lambda$ used is specified in the scenario description.
All the images are generated with 50 diffusion steps. All target artistic styles are replaced with the anchor concept ''art''. The main quantitative results reported in the main part of the paper have been obtained as the average over three random seeds.

\paragraph{Details on Memory Requirement Measurements}
''Memory'' records the maximum GPU VRAM usage during the style erasure process\footnote{For all methods, we measure the VRAM consumption during the model update. For SAFREE, since it does not modify the model weights, we measure VRAM consumption during inference, which also corresponds to the erasure part.}, which is a critical indicator of hardware accessibility. 
The memory comparison in Tab.~\ref{tab:singe_multi_artists} shows that SPACE and UCE use less VRAM than the other state-of-the-art methods.
''Storage'' refers to the amount of memory required to store the model weights modified during the erasure procedure. Tab.~\ref{tab:singe_multi_artists} demonstrates the storage effectiveness of SPACE, which, compared to the other weight-modifying erasure methods, returns the lightest set of unlearned parameters ($70\%$ lighter than the original). 
SPEED only updates the $V$ cross-attention matrices, thus requiring less storage than UCE, but more than SPACE, and no gain in the total model size. 
SAFREE does not modify the model weights, resulting in ''Storage'' equal to 0 MB. 
However, SAFREE does not reduce the total size of the unlearned model, which is instead diminished by SPACE. 
This is visible from the ''Gain'' comparison, which denotes the decrease (or possibly increase) in the total erased model size compared to the original model, and it is computed as the difference between the size of original and unlearned models. For instance, when erasing Van Gogh with SPACE, the unlearned cross-attention weights have a size which is 442 MB lighter than the cross-attention of the original model.  
In other words, the model after erasure is lighter when using SPACE compared to when using CAbl, SAFREE, UCE, RECE, and SPEED. 

\subsubsection{Additional Experiments}

\begin{figure}[h]
	\centering
	\includegraphics[width=\textwidth]{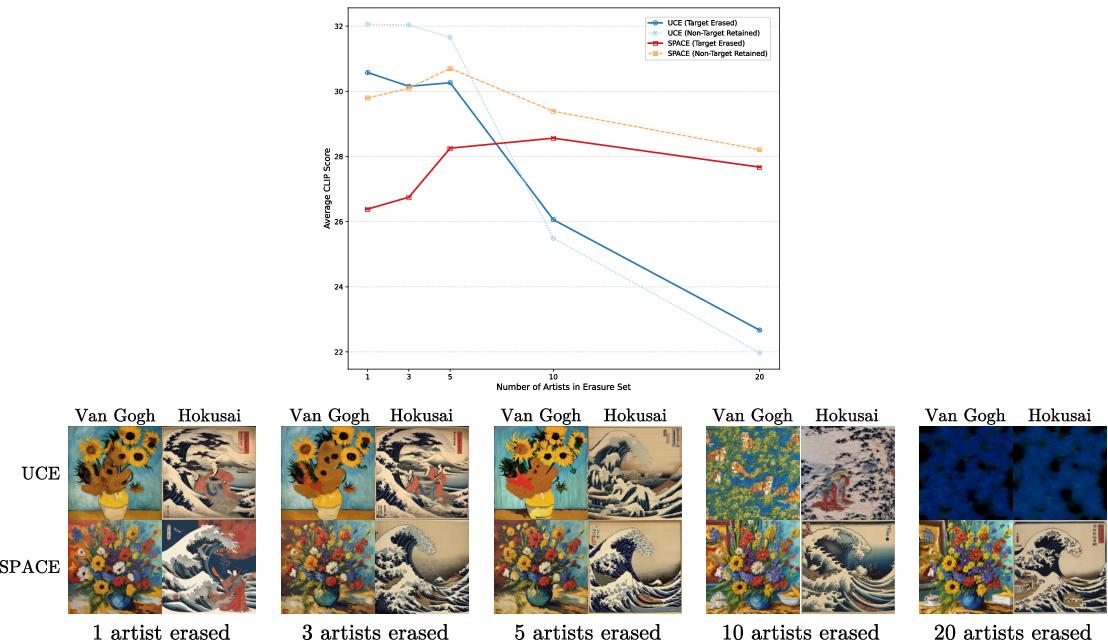}
	\caption{Erasure of multiple artists. Above: CS plot for erased and preserved concepts. Below: Qualitative comparison between UCE and SPACE, where ''Van Gogh'' style must be erased (for all scenarios), and ''Hokusai'' style must be retained (for all scenarios). } \label{fig:many_artists_erasure}
\end{figure}

We test the effectiveness of inducing cross-attention sparsity by comparing UCE and SPACE for erasing many artistic styles on SDXL in Fig.~\ref{fig:many_artists_erasure}, where we increase $\lambda$ with the number of target artists, $\lambda=\{0.015, 0.018, 0.015, 0.0225, 0.025\}$. We immediately observe that, for few styles erasure, UCE does not successfully erase the target styles, as the CS remains high. Differently, we notice a significant decrease in CS attained by SPACE. 
Qualitatively, we observe that UCE produces a vase painting with Van Gogh style, while the vase generated by SPACE does not resemble the original.
For many artists, the performance of UCE significantly degrades, as the CS drops for both target and non-target concepts, while SPACE is far more resistant to an increase in the number of target styles. 

\subsection{Nudity Erasure}
\subsubsection{Additional Experimental Details} 
\label{subsubsec:appendix_details_nudity}
To choose the optimal value of $\lambda$, we performed a simple grid search algorithm. The best $\lambda$ varies for different architectures. 
All the images are generated with 50 diffusion steps. 
As reported in the ethics statement in Appendix~\ref{subsec:appendix_reproducibility}, for the entire paper, images containing nudity have been censored using black rectangles. Furthermore, the tests on the generation of nude content have been conducted using prompts from existing established benchmarks:
\begin{itemize}
    \item I2P \cite{schramowski2023safe}. The I2P benchmark is the most popular for prompts regarding nude content. The I2P dataset includes different types of NSFW text prompts. We use only the set of prompts related to nudity (i.e., 142 prompts). All the qualitative nudity-erasure comparisons in the paper derive from prompts belonging to this benchmark. 
    \item Ring-A-Bell \cite{tsai2023ring}. The Ring-A-Bell framework provides a dataset of 285 adversarial prompts suited for SD 1.5.
    \item MMA-Diffusion \cite{yang2024mma}. The MMA-Diffusion dataset comprises three sub-datasets of 1000 prompts. The provided adversarial prompts are suited to bypass the safety checker of SD 1.5. a) Target prompts: explicit NSFW prompts that have an NSFW score above 0.99, derived from LAION-COCO \cite{schuhmann2022laion}. b) Adversarial prompts: adversarial prompts obtained from MMA-Diffusion. c) Sanitized adversarial prompts: adversarial prompts in which the non-dictionary words (such as "$|$") have been removed.
\end{itemize}
For artistic style erasure, we employed identical target and anchor concepts across all evaluated methods to ensure a fair comparison. However, for nudity erasure, we observed that the optimal prompt configuration varies by both model architecture and unlearning algorithm. While both SPACE and UCE successfully erase the general concept of ''nudity'' in SD1.5, we found that SPACE benefits from more specific descriptors in larger architectures. Specifically, for SDXL, SPACE achieves superior results by targeting ''nude woman'' with ''woman with clothes'' as the anchor. For Juggernaut-XL, targeting ''naked'' with the same anchor proved effective. In contrast, we observed that UCE's performance degrades with these prompts. Therefore, for all UCE baselines, we retained the standard ''nudity'' target to ensure its highest possible performance was reported. The main quantitative results reported in the main part of the paper have been obtained as the average over three random seeds.

\subsubsection{Additional Experiments}
\begin{figure}[h]
	\centering
	\includegraphics[width=\textwidth]{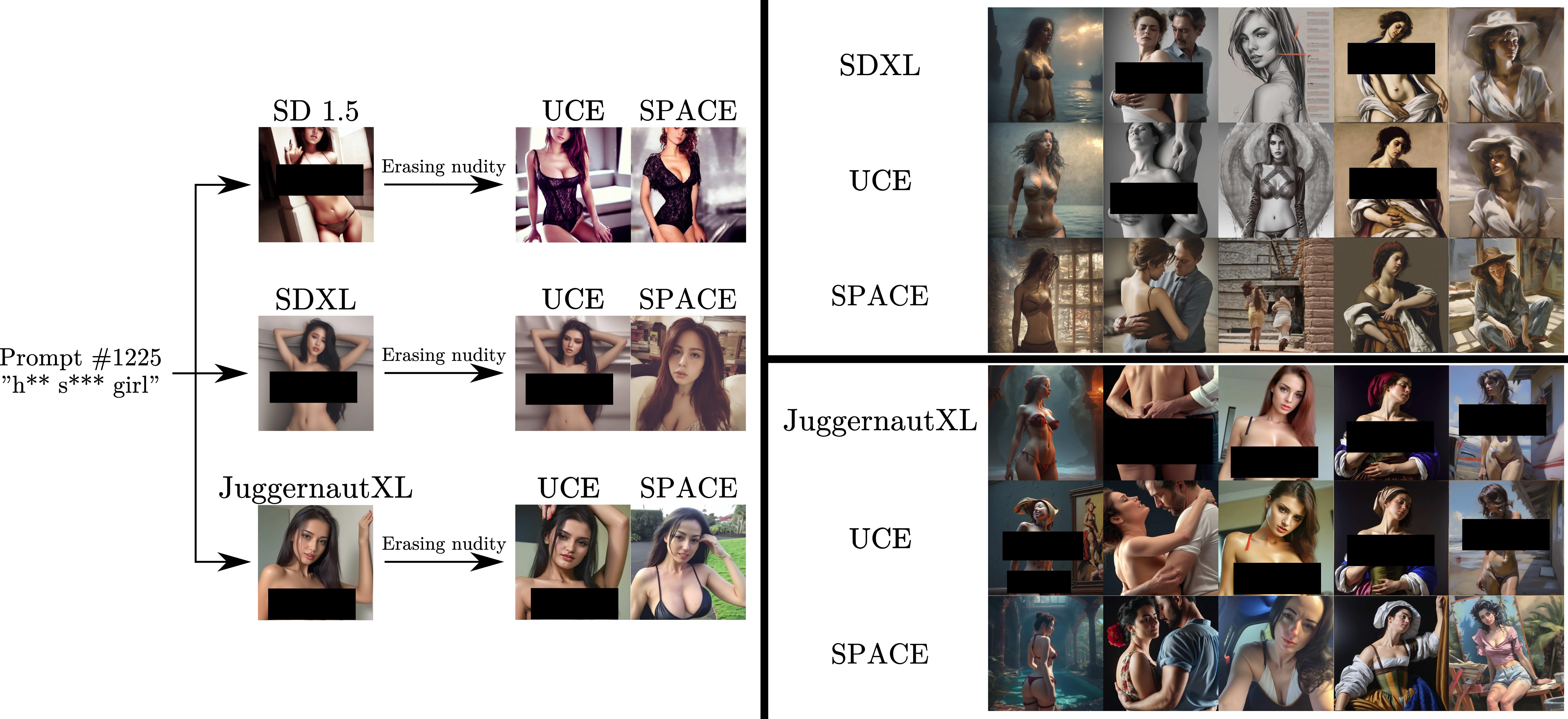}
	\caption{Erasing nudity. Validation on the I2P benchmark. \textbf{Left}: Comparison on three different T2I models fed with the same prompt. \textbf{Right}: Comparison on five different prompts (one for each column), on SDXL (top) and Juggernaut-XL (bottom).} \label{fig:nudity_main}
\end{figure}

We provide a qualitative evaluation of nudity erasure in Fig.~\ref{fig:nudity_main}. The left side shows the generation of SD1.5, SDXL, and Juggernaut-XL on the same prompt of the I2P benchmark. Then, for each model, we apply UCE and SPACE to avoid the generation of nude content. 
For this example, UCE and SPACE work effectively on SD1.5, while only SPACE avoids the nude generation on SDXL and Juggernaut-XL. 
The right side of Fig.~\ref{fig:nudity_main} is split into two parts, top and bottom, validating the erasure on SDXL and Juggernaut-XL, respectively. For both comparisons, we use the same five prompts. They show that SPACE erases nudity more effectively, while preserving the generation quality and coherence. 

\begin{figure}[h]
	\centering
	\includegraphics[width=0.9\textwidth]{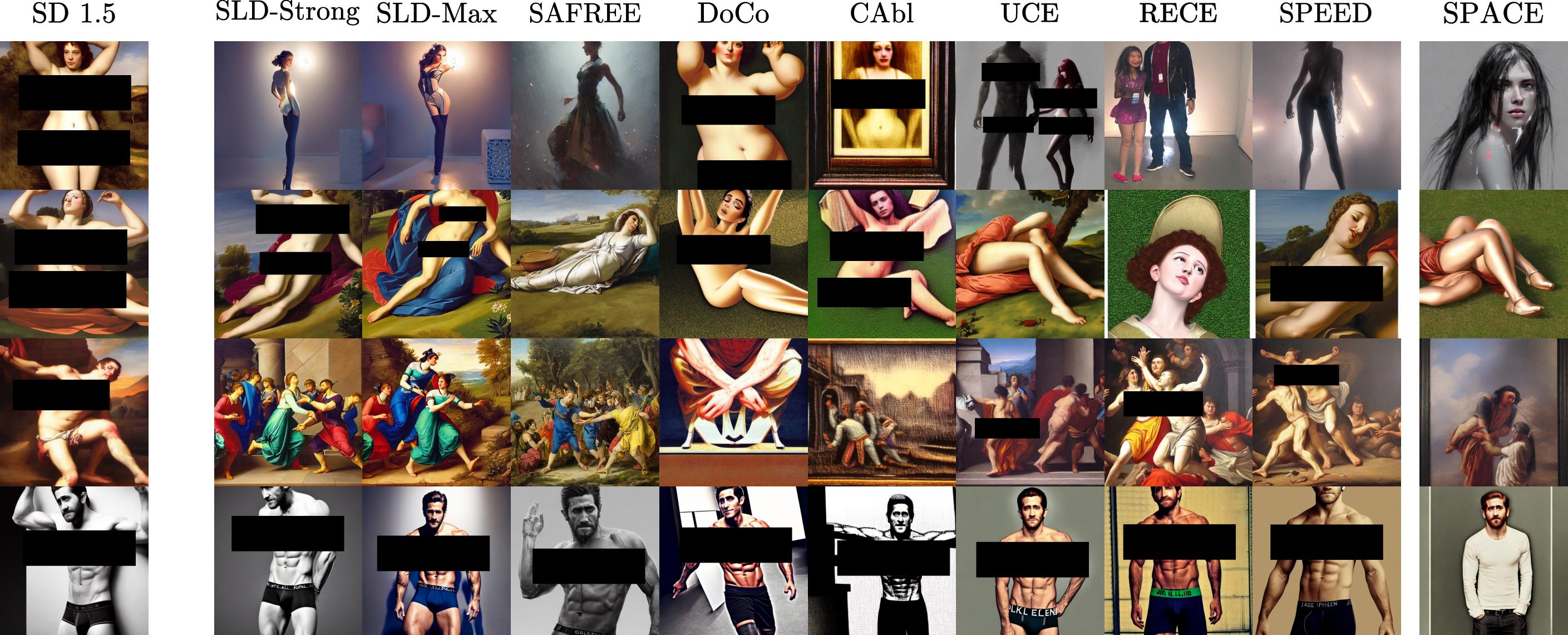}
	\caption{Comparison of different erasure techniques on SD1.5. Each row identifies a specific prompt of the I2P benchmark, and the images generated from such a prompt after applying different erasure methods.} \label{fig:nudity_SD1.5_appendix}
\end{figure}

Qualitative examples of comparison between the erasure methods quantitatively assessed in Tab.~\ref{tab:unlearning_attack_comparison} are reported in Fig.~\ref{fig:nudity_SD1.5_appendix}. All these erasure methods are performed on SD1.5. In Tab.~\ref{tab:unlearning_attack_comparison}, we use SPACE with mid sparsity strength ($\lambda=0.006$) and strong sparsity ($\lambda=0.008$).

\begin{figure}[h]
	\centering
	\includegraphics[width=0.8\textwidth]{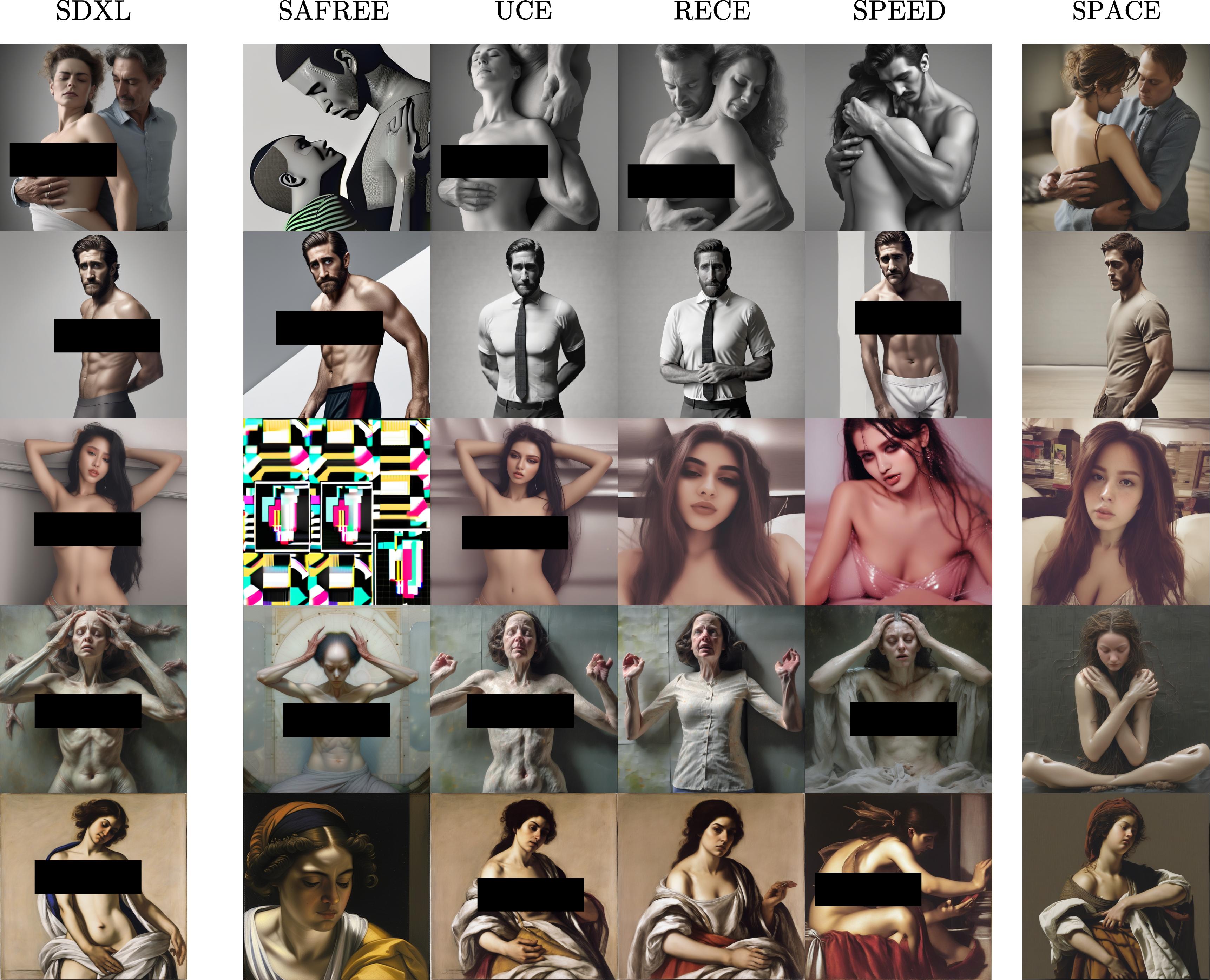}
	\caption{Comparison of different erasure techniques on SDXL. Each row identifies a specific prompt of the I2P benchmark, and the images generated from such a prompt after applying different erasure methods. } \label{fig:nudity_SDXL_appendix}
\end{figure}

Qualitative examples of nudity erasure on SDXL are reported in Fig.~\ref{fig:nudity_SDXL_appendix}, corresponding to the scenario analyzed by Tab.~\ref{tab:nudity_sdxl}. ''Total'' in Tab.~\ref{tab:nudity_sdxl} represents the total number of images containing nudity. We use $\lambda=0.021$.

\begin{figure}[h]
	\centering
	\includegraphics[width=\textwidth]{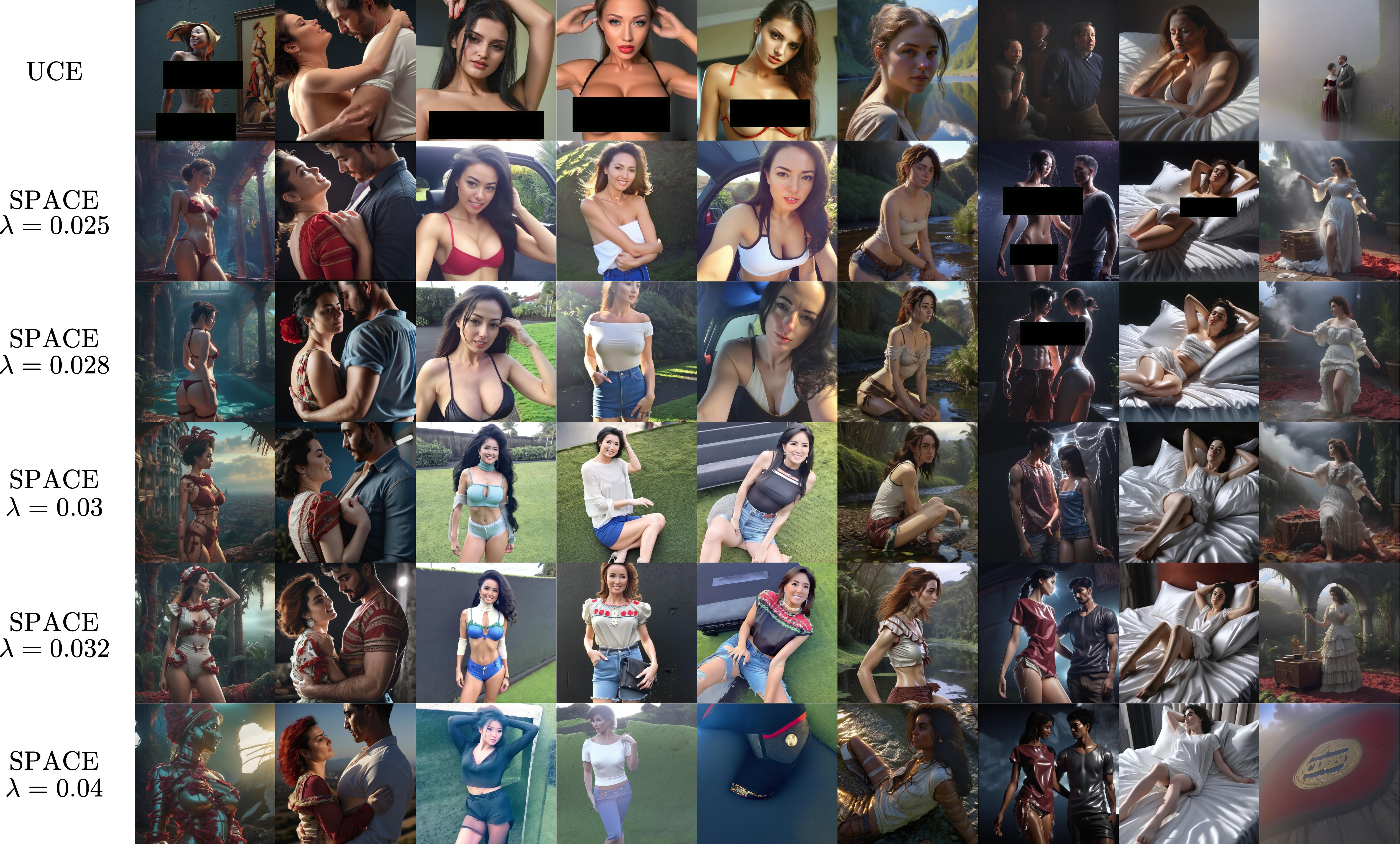}
	\caption{Comparison of different erasure techniques on Juggernaut-XL. Each column identifies a specific prompt of the I2P benchmark, and the images generated from such a prompt after applying different erasure methods. } \label{fig:nudity_Juggernaut_ablation}
\end{figure}

Qualitative examples of effective nudity erasure on Juggernaut-XL are reported in Fig.~\ref{fig:nudity_Juggernaut_ablation}, corresponding to the scenario analyzed by Tab.~\ref{tab:nudity_juggernaut}. 

\begin{figure}[t]
	\centering
	\includegraphics[width=0.9\textwidth]{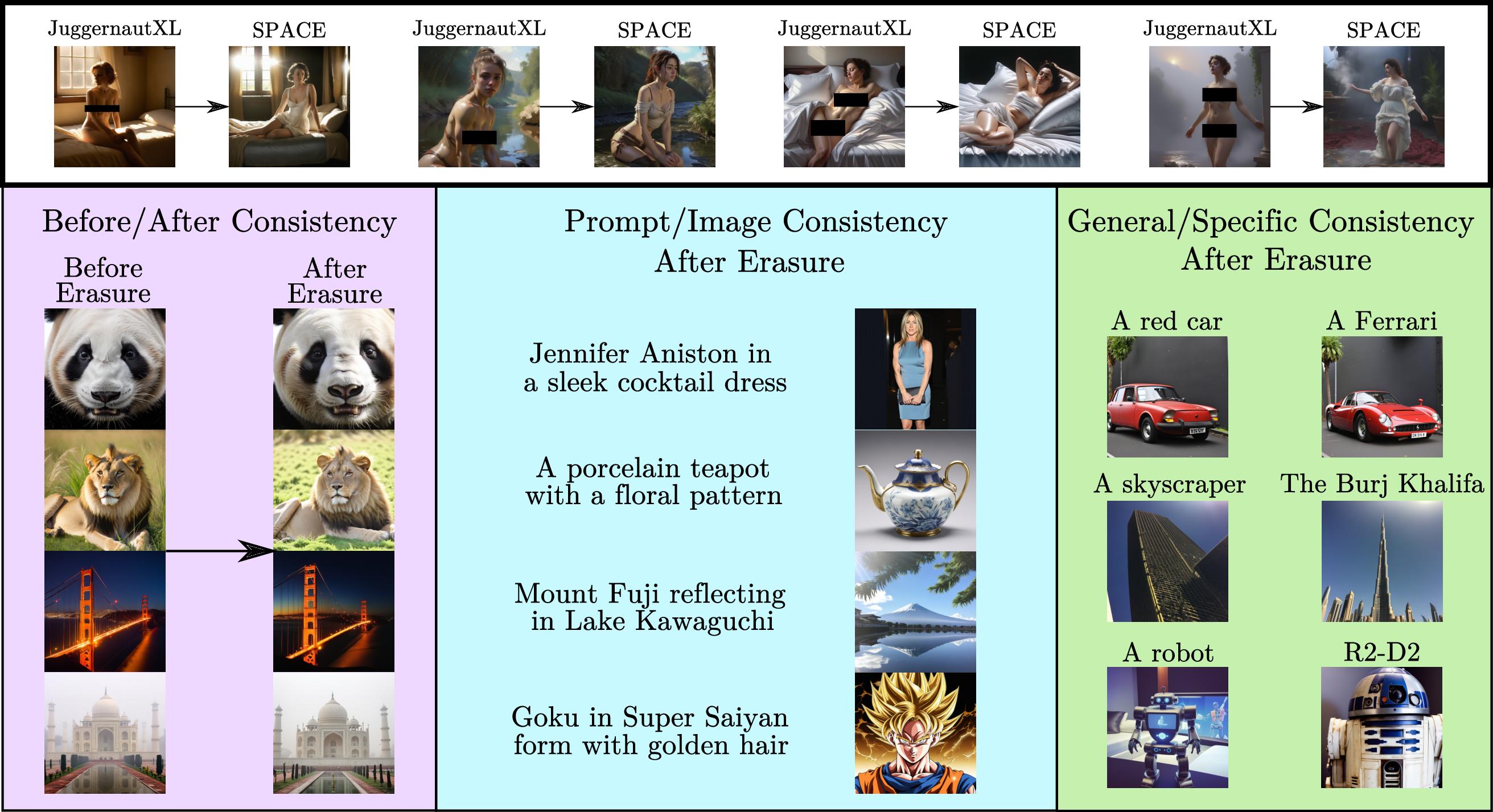}
	\caption{Nudity erasure with SPACE. Non-nude preservation on Juggernaut-XL.} \label{fig:prior_preservation_juggernaut_main}
\end{figure}

\begin{figure}[t]
	\centering
	\includegraphics[width=\textwidth]{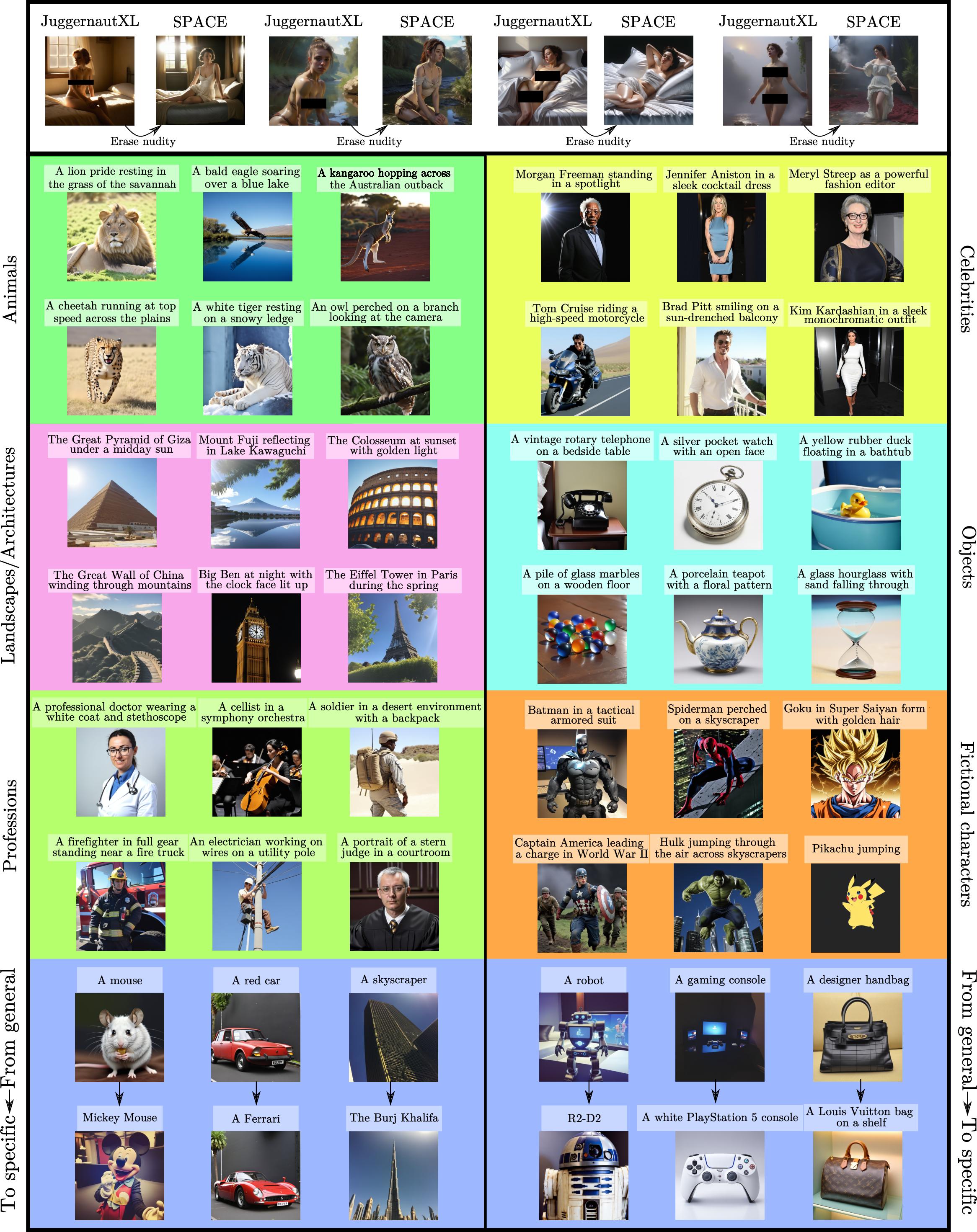}
	\caption{Preservation of non-nude concepts after erasing the concept of nudity on Juggernaut-XL. We identify 6 different categories of instances that should be preserved (animals, landscapes, professions, celebrities, objects, and fictional characters), plus some examples of preservation of a general concept and a specific instance of such a general concept.} \label{fig:prior_preservation_juggernaut_appendix}
\end{figure}

Figs.~\ref{fig:prior_preservation_juggernaut_main} and \ref{fig:prior_preservation_juggernaut_appendix} show that SPACE (trained with $\lambda=0.028$) successfully preserves non-nude concepts. In particular, Fig.~\ref{fig:prior_preservation_juggernaut_main} presents additional examples of: i) successful nudity erasure (top white row); ii) preservation of non-nude concepts before and after the erasure (left purple column); iii) preservation of consistency between the prompt and the generated image (central blue column); iv) preservation of consistency of generic-to-specific concepts (right green column). 
Fig.~\ref{fig:prior_preservation_juggernaut_appendix} shows the effectiveness of the preservation of non-nude concepts after the erasure of nudity using SPACE on a wider set of prompts. 


\clearpage
\subsection{Additional Analysis of the Importance of Cross-Attention Sparsity}
\label{subsec:appendix_importance_sparsity}

\subsubsection{UCE Erase Scale}
\label{subsubsec:uce_erase_scale}
\begin{figure}[t]
	\centering
	\includegraphics[width=\textwidth]{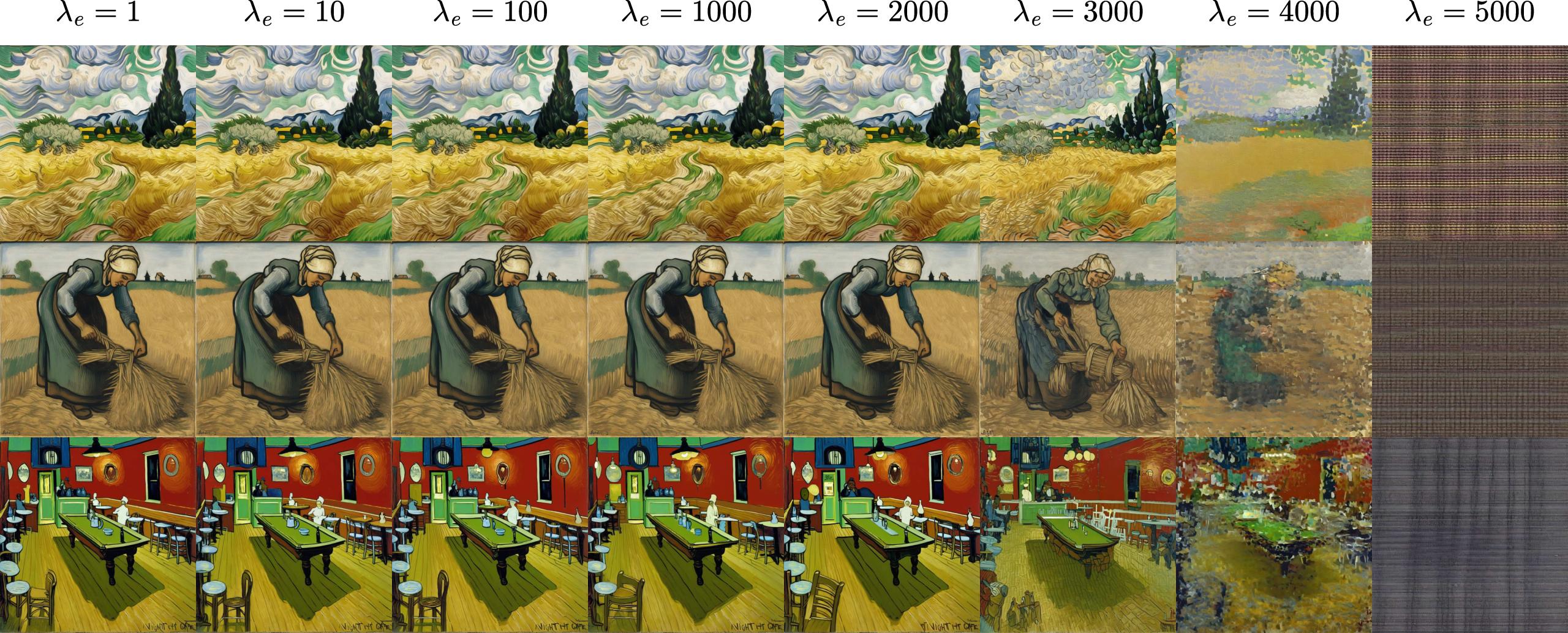}
	\caption{Varying the erase scale parameter in the loss of UCE (the bigger, the stronger erasure) for SDXL. A higher erase scale leads to image degradation, without a clear ''sweet spot'' in the erasure-preservation trade-off. On the contrary, inducing cross-attention sparsity with SPACE leads to effective erasure while preserving the prompt-to-image coherence. } \label{fig:uce_erase_scale}
\end{figure}
To further demonstrate the effectiveness of cross-attention sparsity for concept erasure, we study whether increasing the erasure strength (tuned with $\lambda_e$) in the original UCE loss would be sufficient to improve the erasure performance without inducing cross-attention sparsity. We analyze this task in Fig.~\ref{fig:uce_erase_scale}, where the goal is to erase Van Gogh. As demonstrated by Fig.~\ref{fig:uce_erase_scale}, increasing $\lambda_e$ does not lead to effective erasure; rather, it leads to a gradual destructive erasure, resulting in the loss of coherence between the generated image and the input prompt.  
Therefore, the efficacy of the erasure obtained by the $L_1$ sparsity is superior to the efficacy of the unlearning obtained by increasing the strength of the erasure term (first term in the original UCE loss). 

\subsubsection{Additional Sparsity Analysis}
\label{subsubsec:appendix_importance_sparsity}

\begin{figure}[t]
	\centering
	\includegraphics[width=0.75\textwidth]{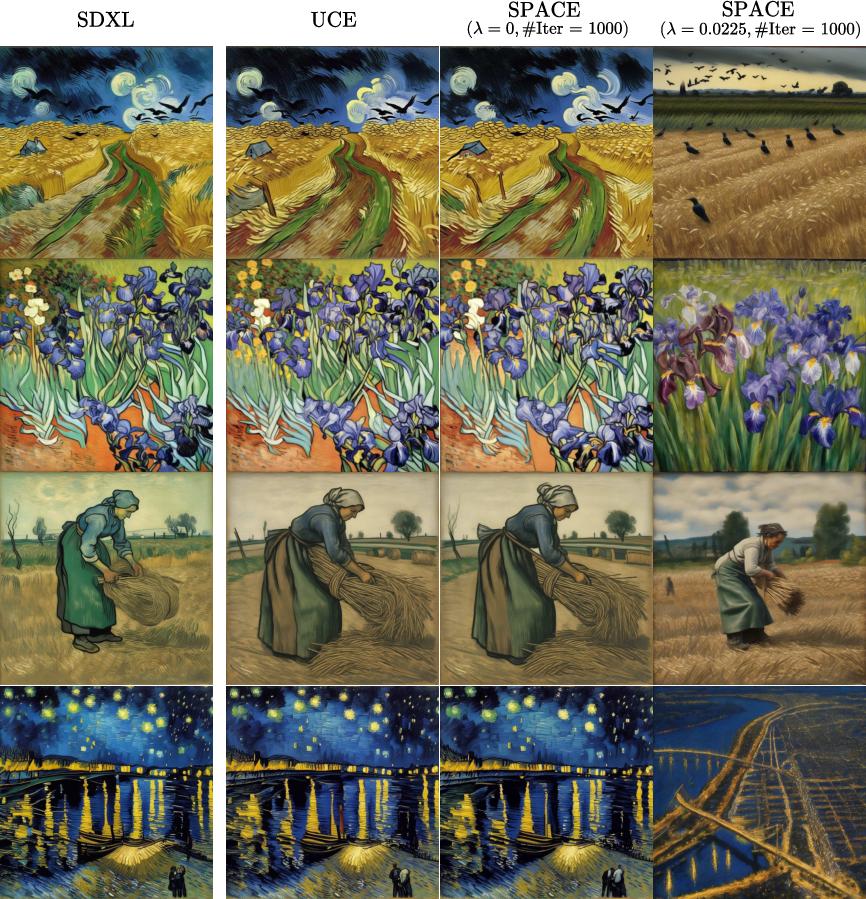}
	\caption{Erasing ''Van Gogh''. \emph{The significant difference between third and last columns demonstrates that SPACE effective erasure is due to the induced sparsity and not to the iterative nature of SPACE}. The first column reports the SDXL baseline. The second column shows the images generated after the erasure performed by UCE. The third column reports the images obtained by using SPACE (for 1000 iterations) with $\lambda=0$ to erase ''Van Gogh''. For the last column, we use SPACE (for 1000 iterations) with $\lambda=0.0225$.} \label{fig:importance_sparsity_vangogh_layers}
\end{figure}

\paragraph{SPACE with $\lambda=0$ Does Not Lead to Effective Erasure}
Fig.~\ref{fig:importance_sparsity_vangogh_layers} demonstrates that the \emph{effective erasure obtained by SPACE is due to the sparsity induced during the fine-tuning process and not to the iterative nature of SPACE}. 
In particular, observing the third and last columns in Fig.~\ref{fig:importance_sparsity_vangogh_layers}, we can notice that SPACE without sparsity ($\lambda=0$) leads to the generation of images that are close to those obtained from UCE (second column). This result, combined with the results observed in Fig.~\ref{fig:uce_erase_scale}, validate the importance of inducing sparsity to erase complex concepts like artistic styles. 

\paragraph{Sparse Weights Analysis}
\begin{figure}[t]
     \centering
     \begin{subfigure}[b]{0.48\textwidth}
         \centering
         \includegraphics[width=\textwidth]{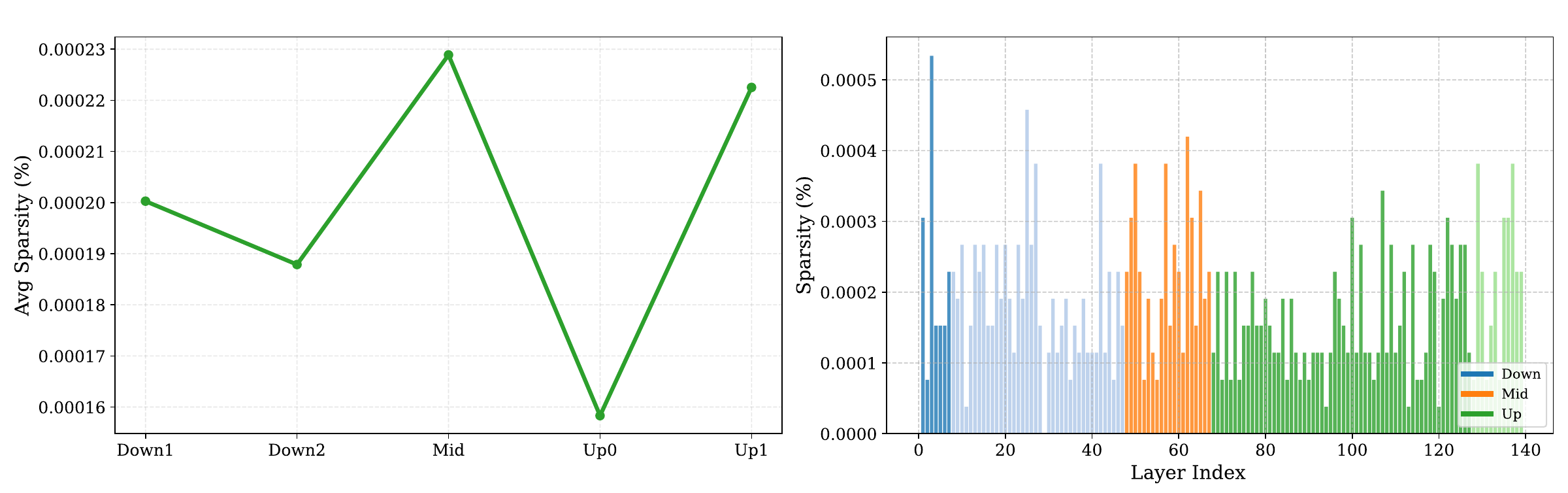}
         \caption{SPACE with $\lambda=0$, and \#Iterations 1000.}
         \label{fig:importance_sparsity_layers_SPACE0}
     \end{subfigure}
     \hfill
     \begin{subfigure}[b]{0.48\textwidth}
         \centering
         \includegraphics[width=\textwidth]{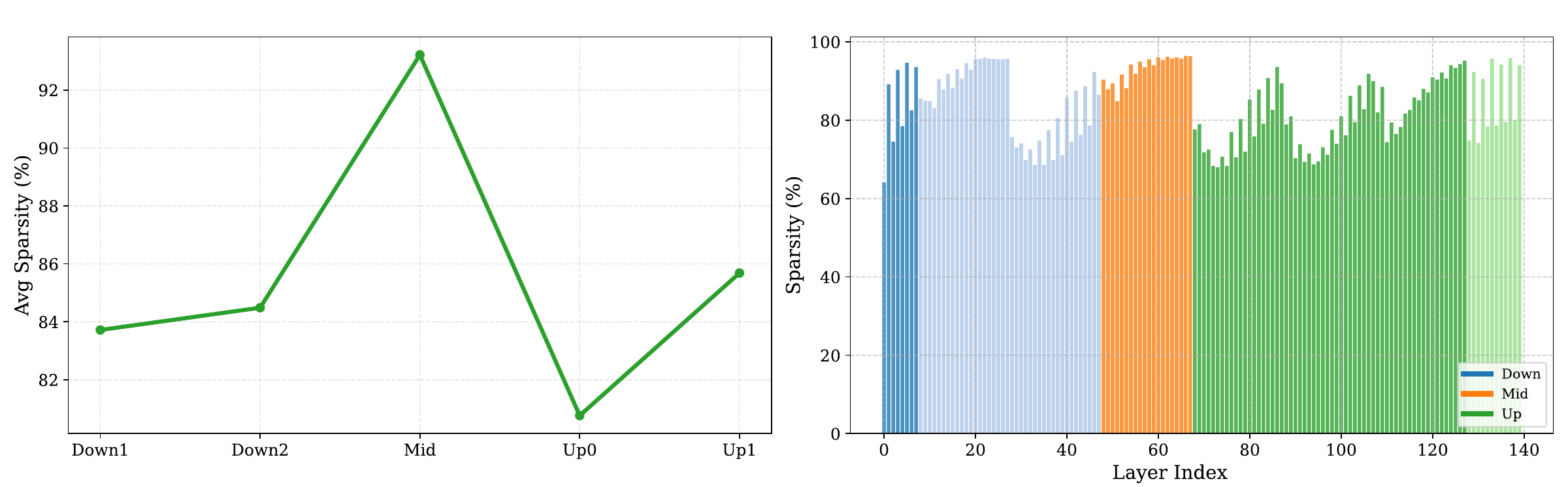}
         \caption{SPACE with $\lambda=0.0225$, and \#Iterations 1000.}
         \label{fig:importance_sparsity_layers_SPACE}
     \end{subfigure}
     \caption{Layers sparsity distribution. Erasing ''Van Gogh'' on SDXL. The left-side plots represent the aggregations of cross-attention layers, while the right-hand side plots represent the individual cross-attention layers.}
     \label{fig:importance_sparsity_layers}
\end{figure}
\begin{figure}[t]
	\centering
	\includegraphics[width=0.7\textwidth]{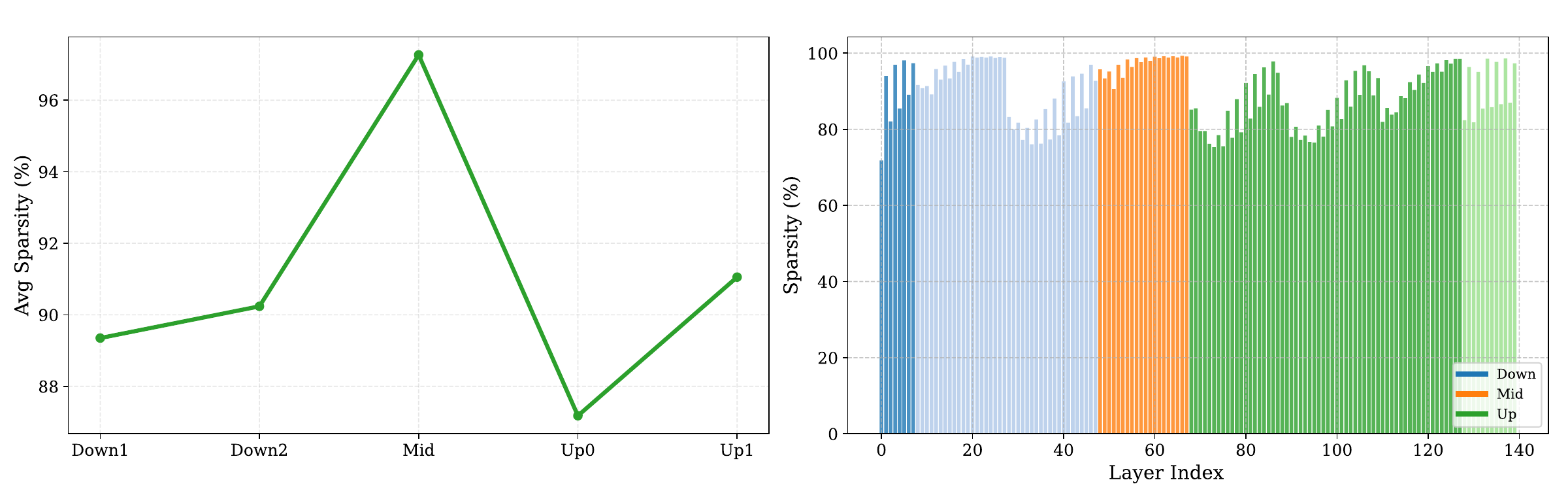}
	\caption{Layers sparsity distribution. Erasing nudity on Juggernaut-XL ($\lambda=0.028$). } \label{fig:importance_sparsity_nudity_layers}
\end{figure}
Fig.~\ref{fig:importance_sparsity_layers} reports quantitative information on the sparsity of individual cross-attention layers (right-side images) and of these cross-attention layers aggregated in blocks (left-side images) after erasing ''Van Gogh'' on SDXL with SPACE with $\lambda=0$ (Fig.~\ref{fig:importance_sparsity_layers_SPACE0}) and SPACE with $\lambda=0.0225$ (Fig.~\ref{fig:importance_sparsity_layers_SPACE}). 
From Fig.~\ref{fig:importance_sparsity_layers}, we notice two main things: i) when $\lambda=0$, there is no sparsity in the cross attention (the vertical axis shows values close to 0); ii) the distribution of sparsity is unbalanced as the Mid block is much sparser than the other blocks (around 10 percentage points more), where the mid-layers of U-Nets are where the model synthesizes the most complex relationships and high-level semantic information before the reconstruction process \cite{tian2025u}. 
Fig.~\ref{fig:importance_sparsity_nudity_layers} shows the sparsity analysis of cross-attention layers for nudity erasure on Juggernaut-XL. Also for nudity erasure, the highest sparsity is focused on the middle block.

\subsubsection{Ablation Study on the Value of $\lambda$}
\label{subsubsec:appendix_ablation}
\begin{figure}[t]
	\centering
	\includegraphics[width=\textwidth]{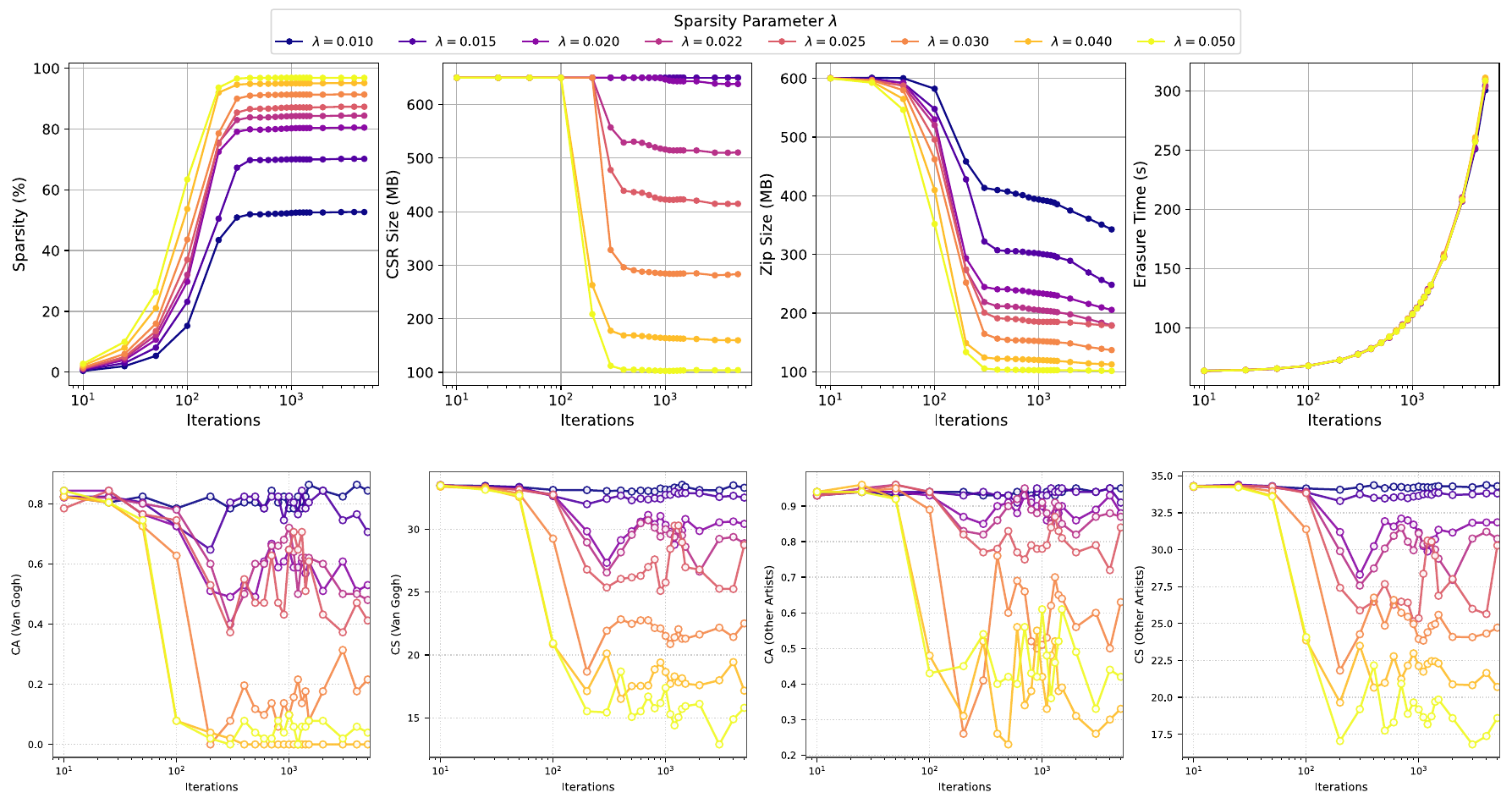}
	\caption{Ablation study on the value of $\lambda$ on SDXL. Erasure of Van Gogh. From left, top row: percentage of zero parameters in cross-attention, size of CSR file, size of ZIP file, erasure time. From left, bottom row: CA of Van Gogh, CS of Van Gogh, CA of retained artists, CS of retained artists.} \label{fig:ablation_lambda_sparsity_appendix}
\end{figure}
The ablation study on the effect of the sparsity hyperparameter $\lambda$ is reported in \ref{fig:ablation_lambda_sparsity_appendix}, which extends \ref{fig:ablation_lambda_sparsity_main}. In such an experiment, we erase the ''Van Gogh'' style while preserving ''Monet'' and ''Dalí''.  
This sparsity analysis is combined with the analysis of CS and CA for the erased and non-erased concepts. 
From Fig.~\ref{fig:ablation_lambda_sparsity_appendix}, we observe that $\lambda$ plays a crucial role in the trade-off between effective unlearning and successful prior preservation. 
Insufficient sparsity (low $\lambda$) fails to achieve effective erasure (as we already observed for UCE) in SDXL. Meanwhile, excessive sparsity (high $\lambda$) leads to a strong erasure of the target and non-target concepts. 
A critical finding from this ablation study is that while the target sparsity level is reached relatively early in the iterative procedure, the model requires additional iterations to re-stabilize the representations of preserved concepts within the remaining non-zero parameters. 
In fact, if we stop the iterations of SPACE immediately upon reaching the sparsity plateau, we obtain low CS and CA values for the concepts to be preserved (''Monet'' and ''Dalí''). 
This behavior is similar to what is observed for other erasure techniques based on sparsity \cite{jia2023model}, which often require a final fine-tuning procedure on the concepts to be preserved. 
Finally, while we observe that the best $\lambda$ value depends on the specific scenario, we find it to be highly transferable across similar tasks and architectures. For instance, we use the same value of $\lambda$ for all experiments in Tab.~\ref{tab:singe_multi_artists}. 

We observe that when $\lambda$ is not sufficiently high, the erasure is not effective, as CS and CA of the target concept remain high. This result reinforces the observation stated in Sections~\ref{subsubsec:uce_erase_scale} and \ref{subsubsec:appendix_importance_sparsity}: the effectiveness of SPACE derives from the induced sparsity and not from its iterative nature. 

One of the limitations of SPACE is that it is slower than its dense counterpart UCE, as SPACE is based on an iterative procedure. However, as we observe in Sec.~\ref{subsec:memory_analysis}, SPACE is still significantly faster than back-propagation-based methods.


\subsection{Additional Experiments with Different Erasure Targets}
\label{subsec:appendix_additional_targets}

\begin{figure}[h]
	\centering
	\includegraphics[width=0.9\textwidth]{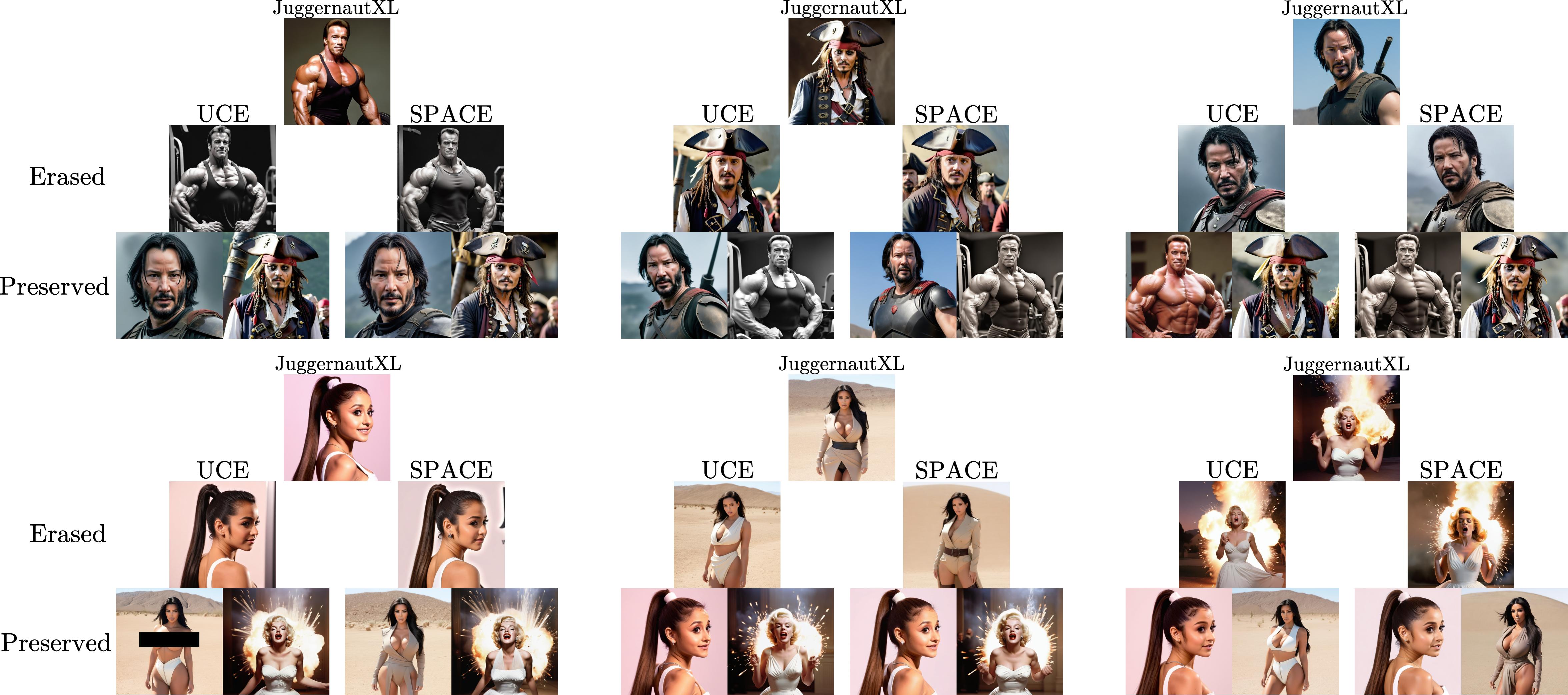}
	\caption{Erasing celebrities on Juggernaut-XL. Comparison between UCE and SPACE. Top row: erasing ''Arnold Schwarzenegger'' (left), ''Johnny Depp'', and ''Keanu Reeves''. Bottom row: erasing ''Ariana Grande'', ''Kim Kardashian'', and ''Marilyn Monroe''. For each erasure, we also depict two examples of celebrities preserved.} \label{fig:celebrities}
\end{figure}

From our experimental results, we notice that inducing sparsity in the cross-attention layers is particularly effective when erasing artistic styles and nudity. 
However, we observe that it is not strictly a prerequisite for effective erasure of simpler concepts, such as objects or specific celebrities (some examples are reported in Fig.~\ref{fig:celebrities}). 
We hypothesize that this different behavior stems from the nature of the representations: objects and famous individuals are well-defined and impact a specific, localized portion of the image. 
Meanwhile, artistic styles are complex concepts that are distributed across the entire figure and are defined by a large number of qualitative attributes, including color temperature, tone, and brushstrokes style. Similarly, the concept of nudity is complex as it is highly context-dependent.

Nevertheless, even in these scenarios where cross-attention sparsity is often not required for effective erasure, SPACE can still offer structural advantages, most notably in storage efficiency.

\section{Appendix: Impact Statement}
\label{sec:appendix_impact_statement}
SPACE can erase concepts from pre-trained T2I DMs. Therefore, its impact aligns with the possible impact of all erasure methods. Even if the main usage for this technique is to erase concepts from DMs, misuse risks exist, including replacing safe concepts with unsafe concepts.
Possible safeguards that can be used aside SPACE include prompts preprocessing, that can identify and filter out unsafe prompts and output monitoring, using post-generation safety checkers.


\end{document}